\documentclass[11pt]{article}

\usepackage[preprint]{acl}

\usepackage{times}
\usepackage{latexsym}

\usepackage[T1]{fontenc}

\usepackage[utf8]{inputenc}

\usepackage{microtype}

\usepackage{inconsolata}

\usepackage{graphicx}
\usepackage{tcolorbox}
\tcbuselibrary{listings, breakable}
\usepackage[T1]{fontenc}
\usepackage{textcomp}
\usepackage{amsmath}
\usepackage{booktabs} 
\usepackage{array}    
\usepackage{tabularx}
\usepackage{multirow}
\usepackage{cuted}

\title{TriQua: Reconciling Granularity and Context in Factuality Evaluation}

\author{
Jin Liu$^{1,2}$ \quad Steffen Thoma$^{1}$ \quad Achim Rettinger$^{1,3}$\\
$^{1}$FZI Research Center for Information Technology, Karlsruhe, Germany \\
$^{2}$Karlsruhe Institute of Technology, Karlsruhe, Germany \\
$^{3}$Trier University, Trier, Germany \\
\texttt{\{jin.liu, thoma, rettinger\}@fzi.de}
}

\begin{document}
\maketitle
\begin{abstract}
The "decompose-then-verify" paradigm for LLM factuality evaluation faces a fundamental trade-off: atomic facts, i.e., one sentence conveying one unit of information, often omit essential context, while broader statements lack the granularity needed for precise assessment. To address this, we introduce TriQua, a framework that flexibly models facts based on their complexity. Simple claims are extracted as standard triples, while complex claims are represented as hyperrelational facts by attaching auxiliary contextual qualifiers. This adaptive structure preserves the necessary context for accurate retrieval and verification without sacrificing atomicity. Furthermore, TriQua's verification process directly annotates concrete errors within specific triples and qualifiers, providing fine-grained explainability for error detection. Alongside the framework, we propose TriQuaScore to quantify the factuality of these structured fact units. Empirical evaluations show that TriQuaScore strongly aligns with human-annotated factuality scores, TriQua achieves robust decomposition quality, and outperforms existing decomposition-based frameworks in evidence-based fact verification.
\end{abstract}

\section{Introduction}

\label{Introduction}
Large language models (LLMs) are increasingly used in real-world applications. However, their tendency to hallucinate factually incorrect or unsupported remains a major barrier to reliable deployment. This has made hallucination detection and factuality evaluation critical for building trustworthy AI systems. The dominant paradigm for factuality evaluation, especially for long-form text, is the "decompose-then-verify" approach, where a complex statement is broken down into smaller, more manageable units for verification. \citet{min-etal-2023-factscore} proposed "atomic fact", a short sentence conveying one piece of information, as the minimal unit to achieve fine-grained analysis. Similarly, \citet{hu-etal-2024-knowledge} decomposed target text into knowledge triplets, formulated as $(\text{subject},\text{relation},\text{object})$. While atomic representations work well for simple, universal facts (e.g., "France is located in Europe."), they struggle with complex assertions. For instance, the statement "As of 2026, France has a population of 64.5 million, excluding overseas departments." \citep{wikidata_qualifiers,enwiki:france_population} contains a core assertion modified by both temporal and geographical constraints (see Table~\ref{tab:extraction_comparison}). Recent work from \citet{song-etal-2024-veriscore} partially addresses the problem by decomposing  text into "verifiable claims" that preserve necessary modifiers to maintain the original assertion's integrity. However, bundling  multiple pieces of modifying information into a single claim violates the principle of atomicity.  Consequently, in a precision-based factuality metric, a single incorrect modifier (e.g., the wrong year) would result in a score of 0 for the entire claim, unfairly penalizing the correct assertion. 

\begin{table*}[t]
\centering
\small
\renewcommand{\arraystretch}{1.4}
\begin{tabularx}{\textwidth}{@{} >{\raggedright\arraybackslash}p{3cm} >{\raggedright\arraybackslash}X @{}}
\toprule
\multicolumn{2}{@{}p{\linewidth}@{}}{\textbf{Fact Decomposition for Text:} \textit{France is located in Europe. As of 2026, France has a population of 64.5 million excluding overseas departments.}} \\
\midrule
\textbf{System} & \textbf{Extraction Output} \\
\midrule
RefChecker \newline \textit{(Triples)} & 
    \textbullet~ ("France", "located in", "Europe") \newline
    \textbullet~ ("France", "has population of", "64.5 million") \newline
    \textbullet~ ("France's population of 64.5 million", "recorded in", "2026") \newline
    \textbullet~ ("France's population of 64.5 million", "excludes", "overseas departments") \\
\midrule
VeriScore \newline \textit{(Sentences)} & 
    \hangindent=1em \textbullet~ France is located in Europe. \par
    \hangindent=1em \textbullet~ As of 2026, the population of France, excluding overseas departments was 64.5 million. \\
\midrule

\textbf{TriQua (Ours)} \newline \textbf{\textit{(Triples + Qualifiers)}} & 
    \textbullet~ {\ttfamily \{"base\_triple": ["France", "is located in", "Europe"]\}} \newline
    \textbullet~ {\ttfamily \{"base\_triple": ["France", "has population of", "64.5 million"],} \newline
    \phantom{\textbullet~ }{\ttfamily \hspace{2mm}"qualifiers": [\{"time": "2026"\},} \newline
    \phantom{\textbullet~ }{\ttfamily \hspace{2mm}\phantom{"qualifiers": [}\{"excluding": "overseas departments"\}]\}} \\
\bottomrule
\end{tabularx}
\caption{Comparison of extracted decompositions for text containing both simple and complex claims. While baselines like RefChecker and VeriScore struggle to balance atomicity and context, our TriQua framework flexibly adapts by natively supporting both simple facts (standalone base triples) and complex facts (base triples paired with independent qualifiers). This dual representation prevents unfair penalization and enables explainable verification.}
\label{tab:extraction_comparison}
\end{table*}

This presents a fundamental trade-off: the fine granularity of atomic facts is lost when preserving the broader context of verifiable claims, yet context is often crucial for accurate verification. While prior works, such as \citet{wang-etal-2024-factcheck, metropolitansky-larson-2025-towards}, have addressed the atomicity and context problem, a comprehensive solution has not yet been established. To address this challenge, we draw inspiration from structured knowledge representation and adapt it for Open Information Extraction (OpenIE) over unstructured text. In this paper, we introduce a framework that handles varying textual complexity through two distinct representations: simple facts are extracted purely as atomic $(\text{subject}, \text{relation}, \text{object})$ base triples, while complex facts are represented by pairing a base triple with a set of independent qualifiers (as demonstrated in Table~\ref{tab:extraction_comparison}). The base triple captures the core information similar to an atomic fact, while qualifiers are used to annotate and contextualize this core assertion with the necessary spatial, temporal, causal, or other modifying information. By doing so, our method harmonizes the strengths of both prior approaches, enabling a fact verification framework that is both highly granular and context-aware. 

Building on this decomposition, we propose a relation-based factual precision metric. Because our architecture cleanly separates the base triple from its various qualifiers, our verification process can explicitly annotate which specific component (the base triple or particular qualifier) is incorrect when a factual unit is unsupported. This targeted annotation provides granular explainability and consequently allows the metric to assign partial scores rather than penalizing an entire claim for a single faulty modifier. Ultimately, this yields a higher diagnostic assessment of factuality while retaining all contextual information required for subsequent retrieval and verification.

In summary,our main contributions are as follows:
\begin{itemize}
    \item We propose TriQua, an OpenIE-based decomposition framework that represents simple facts as base triples and complex facts as base triples with qualifiers, preserving both atomicity and contextual constraints.
    \item We introduce TriQuaScore, a relation-based factual precision metric that enables partial scoring and localized error attribution.
    \item We evaluate TriQua against decomposition-based baselines across metric alignment, decomposition quality,  and short-claim fact checking, demonstrating improved factuality scoring and more interpretable verification.
\end{itemize}

\section{Related Work}
\label{related_work}
\textbf{Decomposition and decontextualization:} Factuality evaluation of LLM generations generally relies on the "decompose-then-verify" paradigm, pioneered by FactScore~\citep{min-etal-2023-factscore}. However, a primary challenge of this approach is that extracting strictly atomic facts often strips away essential contextual information. To address these limitations, particularly concerning context preservation and verifiability in open-ended generation, subsequent frameworks such as Factcheck-GPT~\citep{wang-etal-2024-factcheck}, SAFE~\citep{wei2024longform}, VeriScore~\citep{song-etal-2024-veriscore}, Molecular Facts~\citep{gunjal-durrett-2024-molecular}, and DnDScore~\citep{wanner-etal-2025-dndscore} have extended the initial FactScore methodology. Most frameworks represent extracted facts as sentence, while others adopt more structured representations. RefChecker~\citep{hu-etal-2024-knowledge} uses knowledge triples, and QASemConsistency~\citep{cattan-etal-2026-localizing} represents factual content through predicate-argument propositions, expressed as QA pairs, to evaluate factuality of the generated text. Decontextualization also differs across systems. SAFE, Molecular Facts, and DnDScore employ a dedicated decontextualization step following decomposition. In contrast, Factcheck-GPT and VeriScore merge decomposition and decontextualization into a single unified step, which is also adopted by our framework. Furthermore, the specific targets of decontextualization vary across the literature: SAFE focuses on resolving pronouns and definite references; Molecular Facts and DnDScore prioritize entity disambiguation and minimality; and VeriScore targets spatial and temporal modifiers. While VeriScore's \textit{modifier} closely aligns with our proposed \textit{qualifier} concept, our approach is significantly more systematic, highly structured, and designed to cover a much broader spectrum of context types.

\textbf{Decomposition Quality Check:} Several recent studies address the challenges of assessing decomposition quality. Highlighting the pitfalls of this process, \citet{hu-etal-2025-decomposition} classify typical errors, including context omission, ambiguity, over-decomposition, and alteration of the original meaning. To quantitatively measure decomposition success, \citet{wanner-etal-2024-closer} introduced DecomposeScore, which calculates the proportion of subclaims accurately supported by the original text. Expanding on multidimensional evaluation, \citet{metropolitansky-larson-2025-towards} proposed the CLAIMIFY framework to assess decomposition across three dimensions: entailment, decontextualization, and coverage. Similarly, the FactLens benchmark~\citep{mitra-etal-2025-factlens} evaluates metrics such as atomicity, sufficiency, fabrication, and redundancy by comparing them against human annotations.

\textbf{Factuality metrics:} FactScore measures factuality as precision: the proportion of supported facts among all decomposed facts. A limitation of precision-only scoring is that shorter responses can receive inflated scores by making fewer claims. SAFE addresses this by defining a target number of facts $K$ and reporting recall and $F1@K$, an idea later adopted by VeriScore. VeriFact~\citep{liu-etal-2025-verifact} similarly estimates recall using LLM- and human-generated reference facts. Since choosing $K$ or constructing reference facts introduces additional subjectivity, TriQua focuses on factual precision while improving the granularity of the counted fact units.

\textbf{Structured knowledge representation:} TriQua builds on established ideas from structured knowledge representation, including n-ary relations~\citep{n-ary}, Wikidata-style qualifiers~\citep{acm_wikidata}, and hyper-relational facts~\citep{hyperrelationfact_rosso, wei-etal-2024-shot}. These formalisms extend standard triples with additional arguments or qualifier pairs to represent contextual constraints. TriQua adapts this idea to OpenIE-based factuality evaluation of unstructured LLM generations.

\section{Methodology}
In this section, we introduce the components in our proposed TriQua framework for fine-grained factuality evaluation. Our framework also follows the decompose-then-verify paradigm. We first introduce the core components, namely decomposer, retriever and verifier. In addition to the components, we also introduce our own factuality metrics based on the verification results.
\subsection{Decomposition}
\subsubsection{Fact representation}
While most existing decompose-then-verify frameworks represent extracted facts as unstructured sentences, TriQua utilizes a structured approach inspired by the $(\text{subject}, \text{relation}, \text{object})$ triple. However, standard triples struggle to capture complex  contextual details, such as temporal or conditional constraints, that are often critical for accurate verification. To overcome this limitation, TriQua adopts the hyperrelational fact representation~\citep{wei-etal-2024-shot}, which augments a base triple with auxiliary qualifier-value pairs. We define facts as follows:
\begin{itemize}
    \item Simple Facts are represented as standard triples: $(\text{subject}, \text{relation}, \text{object})$
    \item Complex Facts are represented as a base triple modified by a set of $m$ qualifiers: 
$((\text{subject}, \text{relation}, \text{object}), \{(q_{i},v_{i})\}_{i=1}^{m})$, where $q_{i}$ is the $i$-th qualifier label and $v_{i}$ is its corresponding qualifier value.
\end{itemize}
\subsubsection{Qualifier usage}
\label{sec:qualifier_usage}
Following \cite{aljalbout-qualifiers}, we categorize qualifiers into validating context (e.g., temporal or spatial), causality, sequence, provenance, and annotations (e.g., constraints or attributes). Qualifiers are applied only when necessary. If a general entity property does not directly modify the assertion of a base triple, it should be extracted as an independent  simple fact rather than appended as a qualifier. Furthermore, a base triple can have multiple qualifiers. We follow Wikidata's convention to prevent ambiguity: qualifiers can only modify the base triple and not other qualifiers~\citep{acm_wikidata, wikidata_qualifiers}. 
\subsubsection{Few-shot open information extraction}
\label{sec:few_shot}
Our decomposition follows the open information extraction paradigm, which eliminates the need for a fixed predefined ontology. Furthermore, the process is strictly constrained to extracting only factual information. Guided by the aforementioned rules, we utilize few-shot learning to extract objective facts from the target text for subsequent verification. Our framework is highly adaptable, supporting both sentence- and document-level decomposition. For document-level texts, we employ the recursive text splitter from LangChain~\citep{chase2022langchain}. The tool divides the text into non-overlapping chunks of adjustable size, prioritizing natural semantic boundaries by splitting sequentially at paragraphs, lines, and sentences. When decomposing a chunk, we also provide the immediately preceding chunk as context.  The prompt is instantiated according to two input conditions: whether an original user question is available,  and whether chunking is needed. This yields four modes: with/without question and with/without chunking. The question, when available, provides task context for resolving underspecified references. the previous chunk, when available, provides local context. The mode-specific details are shown in Appendix~\ref{sec:decomposition_modes}. Exact prompt templates are provided in Appendix~\ref{sec:decomp_promp_wq_wc} and Appendix~\ref{sec:decomp_promp_woq_woc}.
\subsection{Retrieval}
Since our extraction pipeline relies on open information extraction without a predefined ontology, and we primarily rely on unstructured text for our reference information, we convert our structured fact representations into plain strings. Specifically, for a complex fact formulated as $((\text{subject}, \text{relation}, \text{object}), \{(q_{i},v_{i})\}_{i=1}^{m})$, we concatenate its components into a query string "$\text{subject  relation  object } q_{1} \ v_{1} ... q_{m} \ v_{m}$". For simple facts, the query string reduces to a base triple: "$\text{subject relation object}$". The choice of the specific retriever, such as commercial web search engines (e.g., Google), dense vector retrievers, or LLM-based retrieval systems, depends on the characteristics of the data.   
\subsection{Fact verification}
\label{sec:fact_verification}
Given the decomposed facts and retrieved references, we utilize an LLM to verify each fact using a Natural Language Inference (NLI) format. Depending on its complexity, the fact is formulated for verification as either a simple fact consisting of the base triple "$(subject, relation, object)$" or a complex fact modified by its qualifiers "$(subject, relation, object)-(q_{1},v_{1})-...-(q_{m},v_{m})$". Compared to traditional NLI tasks, where models output binary or ternary labels, we additionally instruct the model to annotate the specific incorrect components when encountering unsupported facts. This approach provides granular explainability alongside the standard verification label. The complete prompt is provided in Appendix~\ref{sec:verification_annotation_prompt}.

\subsection{Factuality metric}
\label{sec:factuality_metric}
Following the verification of decomposed facts from previous steps, the TriQua framework calculates an overall factuality score for the original target text (whether a sentence, paragraph, or document). In this work, we focus specifically on the precision score. 

To ensure a fine-grained evaluation, we establish a strict distinction between a Fact and a Fact Unit. While a fact represents the complete structured claim evaluated in the previous step, a fact unit is the atomic, quantifiable building block used for calculating our metric. We propose a relation-based metric because both triples and qualifiers are fundamentally defined by relations. Specifically, while a base triple centers on a primary relation, the label in qualifiers is itself a relation. Since both components represent a relation, we can break facts down to quantifiable units: \textbf{the base triple constitutes 1 fact unit and each attached qualifier constitutes an additional 1 fact unit}. Therefore, a simple fact is counted as $1$ fact unit, whereas a complex fact modified by $m$ qualifiers is counted as $m+1$ fact units. 

To compute the factuality metric, we isolate the supported fact units from the unsupported fact units using their assigned verification labels (e.g., Contradiction or Neutral) and corresponding error span annotations. First, both the generated fact units and the annotated error spans undergo textual normalization, removing punctuation to prevent artificial mismatches. Next, we employ a substring matching heuristic to map the error spans back to their atomic fact units. A fact unit is flagged as unsupported if it contains the annotated error text span.

To account for cases where duplicate qualifiers or base triples might exist across the text, we calculate the overall \textbf{TriQuaScore} using the \textbf{deduplicated} counts of these units:
\begin{equation}
\text{TriQuaScore}=\frac{|\text{Supported  Fact  Units}|}{|\text{Total Fact Units}|}
\end{equation}

\section{Experiments}
To systematically evaluate TriQua, we compare against VeriScore~\citep{song-etal-2024-veriscore} and RefChecker~\citep{hu-etal-2024-knowledge}, which are the closest decomposition-based baselines: VeriScore preserves modifiers within verifiable claims, while RefChecker uses triple-based fact representations. For transparency and reproducibility, we conduct all experiments with open-weight models, using Qwen3.5-397B-A17B~\citep{qwen3.5} as the primary backbone and including additional Qwen3.5 scales and model families.

Our evaluation focuses first on the end task: how well each framework's factuality metric aligns with human factuality judgments on long-form generations. We then analyze intrinsic decomposition quality, including entailment, decontextualization, and qualifier usage, to understand whether the extracted fact units preserve both atomicity and context. Finally, we report evidence-based binary fact-checking results on CLEARFACTS as a stress test for short, complex claims, where direct verification provides a strong non-decomposition baseline. Full experimental details are provided in Appendix~\ref{sec:experiment_setup}.
\subsection{Factuality metric evaluation}
\label{sec:metric_evaluation}
\begin{table*}[t]
\centering
\scriptsize
\setlength{\tabcolsep}{3pt}

\begin{minipage}[t]{0.55\textwidth}
\centering
\begin{tabular}{llccc}
\toprule
\textbf{Backbone Model} & \textbf{Framework} & \textbf{Pearson ($r$) $\uparrow$} & \textbf{Spearman ($\rho$) $\uparrow$} & \textbf{MAE $\downarrow$} \\
\midrule
\multirow{3}{*}{Qwen3.5-397B-A17B} 
& RefChecker & 0.8917 & \textbf{0.8789} & 0.0847 \\
& VeriScore & 0.8775 & 0.8579 & 0.0895 \\
& \textbf{TriQua} & \textbf{0.8929} & 0.8725 & \textbf{0.0821} \\
\midrule
\multirow{3}{*}{Qwen3.5-122B-A10B} 
& RefChecker & 0.8938 & 0.8805 & 0.0853 \\
& VeriScore & 0.8871 & 0.8642 & 0.0873 \\
& \textbf{TriQua} & \textbf{0.8983} & \textbf{0.8817} & \textbf{0.0824} \\
\midrule
\multirow{3}{*}{Qwen3.5-35B-A3B} 
& RefChecker & \textbf{0.9006} & \textbf{0.8911} & 0.0875\\
& VeriScore & 0.8836 & 0.8625 & 0.0883 \\
& \textbf{TriQua} & 0.8917 & 0.8781 & \textbf{0.0845} \\
\midrule
\multirow{3}{*}{GLM-5-FP8 (744B)} 
& RefChecker & 0.9049 & \textbf{0.8934} & 0.0899 \\
& VeriScore & 0.8905 & 0.8752 & 0.0892 \\
& \textbf{TriQua} & \textbf{0.9053} & 0.8892 & \textbf{0.0806} \\
\midrule
\multirow{3}{*}{GPT-OSS-120B} 
& RefChecker & 0.8554 & 0.8534 & 0.1006 \\
& VeriScore & 0.8882 & 0.8685 & 0.0876 \\
& \textbf{TriQua} & \textbf{0.9080} & \textbf{0.8949} & \textbf{0.0785} \\
\bottomrule
\end{tabular}
\end{minipage}%
\hfill
\begin{minipage}[t]{0.43\textwidth}
\centering
\begin{tabular}{lccc}
\toprule
\textbf{TriQua Variant} & \textbf{Pearson ($r$) $\uparrow$} & \textbf{Spearman ($\rho$) $\uparrow$} & \textbf{MAE $\downarrow$} \\
\midrule
TriQua & \textbf{0.8929} & \textbf{0.8725} & \textbf{0.0821} \\
\quad Claim-level & 0.8817 & 0.8617 & 0.0920 \\
\quad w/o Deduplication & 0.8912 & 0.8681 & 0.0850 \\
\midrule
TriQua & \textbf{0.8983} & \textbf{0.8817} & \textbf{0.0824} \\
\quad Claim-level & 0.8916 & 0.8807 & 0.0931 \\
\quad w/o Deduplication & 0.8897 & 0.8721 & 0.0880 \\
\midrule
TriQua & \textbf{0.8917} & \textbf{0.8781} & \textbf{0.0845} \\
\quad Claim-level & 0.8852 & 0.8714 & 0.1007 \\
\quad w/o Deduplication & 0.8852 & 0.8700 & 0.0880 \\
\midrule
TriQua & \textbf{0.9053} & \textbf{0.8892} & \textbf{0.0806} \\
\quad Claim-level & 0.8823 & 0.8697 & 0.0986 \\
\quad w/o Deduplication & 0.8941 & 0.8735 & 0.0864 \\
\midrule
TriQua  & \textbf{0.9080} & \textbf{0.8949} & \textbf{0.0785} \\
\quad Claim-level & 0.8990  & 0.8874 & 0.0955 \\
\quad w/o Deduplication & 0.9023 & 0.8865 &0.0822  \\
\bottomrule
\end{tabular}
\end{minipage}

\caption{
Main factuality metric evaluation and TriQua ablations across backbone models.
\textbf{Left}: comparison with decomposition-based baselines on FactScore-Bio.
\textbf{Right}: ablation variants of TriQua, aligned with the same backbone order as the left panel.
Claim-level scores each decomposed claim as a whole, while w/o Deduplication counts repeated fact units separately.
Best values within each backbone block are bolded.
}
\label{tab:aligned_main_and_ablation}
\end{table*}

We evaluate our factuality metric using the biographical dataset introduced by FactScore~\cite{min-etal-2023-factscore} (hereafter referred as FactScore-Bio), which contains 157 ChatGPT-generated biographies. Since all prompt subjects have Wikipedia profiles, we scrape and chunk their corresponding articles. We then use Qwen3-Reranker-8B~\citep{qwen3embedding} to retrieve the top-10 most relevant chunks per decomposed claim. The retrieval setup is identical across all frameworks, varying only by the query formulation. TriQua queries are formatted as either "$\text{subject relation object}$" or "$\text{subject  relation  object } q_{1} \ v_{1} ... q_{m} \ v_{m}$". RefChecker queries are constructed by converting triples to a "$\text{subject relation object}$" format, whereas VeriScore directly queries the decomposed sentences. After retrieval, the corresponding verifier evaluates the claims.

Focusing on precision-based factuality, we define the human-annotated gold factscore as the proportion of supported atomic facts against total extracted atomic facts. For the framework-predicted scores, TriQua applies the TriQuaScore  introduced in Section~\ref{sec:factuality_metric}, while VeriScore and RefChecker use the proportion of supported decomposed facts. We evaluate the alignment between human-annotated gold scores and framework predictions using Pearson correlation, Spearman correlation, and Mean Absolute Error (MAE). Table~\ref{tab:aligned_main_and_ablation} left panel details the comparative results across the different backbone models. The data demonstrates a strong alignment between our TriQuaScore and the human-annotated gold scores, with Pearson correlations consistently reaching between 0.89 and 0.90. While RefChecker proves highly competitive regarding the correlation metrics, often matching or slightly outperforming TriQua on Spearman $\rho$, TriQuaScore maintains absolute dominance in error reduction. TriQua consistently achieves the lowest MAE across all five tested backbone models. Specifically, when compared to VeriScore, TriQuaScore yields an MAE reduction of approximately 4-10\%. 

\textbf{Ablation:} We further ablate TriQuaScore with two variants: claim-level scoring and w/o Deduplication. Claim-level scoring treats each extracted simple or complex fact as one claim, i.e., a base triple together with all attached qualifiers, making it conceptually close to VeriScore. The w/o Deduplication variant computes the proportion of supported fact units without merging repeated units. As shown in the right panel of Table~\ref{tab:aligned_main_and_ablation}, the full TriQuaScore consistently outperforms both variants. The gain over claim-level scoring shows the benefit of evaluating base triples and qualifiers as separate fact units. Deduplication further improves robustness to repeated contextual units introduced during decomposition. For example, for ``Ronaldo played for Corinthians, where he won the Campeonato Paulista and the Copa do Brasil.'' TriQua may extract two facts:
\textit{(Ronaldo, won, Campeonato Paulista)-(played for, Corinthians)} and
\textit{(Ronaldo, won, Copa do Brasil)-(played for, Corinthians)}.
Here, the contextual unit \textit{(played for, Corinthians)} appears twice. Without deduplication, the denominator contains four fact units, although the sentence expresses three distinct pieces of information: Ronaldo played for Corinthians, won the Campeonato Paulista, and won the Copa do Brasil. Our deduplication step simply counts repeated fact units once, yielding a score denominator that better matches human annotations.
\subsection{Decomposition quality evaluation}
While decomposition quality can be assessed from multiple perspectives, we evaluate our decomposition quality across three dimensions. In Section~\ref{sec:entailment_decontextualization}, we benchmark TriQua against existing frameworks on entailment and decontextualization. In Section~\ref{sec:qualifier_evaluation}, we conduct an intrinsic evaluation specific to TriQua's hyperrelational structure, assessing the structural and semantic validity of its qualifier usage.
\label{sec:decomposition_quality}
\subsubsection{Entailment and decontextualization}
\label{sec:entailment_decontextualization}
We evaluate decomposition quality with two automatic diagnostic checks. The entailment check tests whether each extracted fact preserves the meaning of the source text without introducing unsupported content. The decontextualization check tests whether the fact is self-contained, i.e., whether references such as pronouns, definite descriptions, relative times/places, or comparative terms are resolved when the source provides their antecedents. Ambiguity inherited from the source itself is not penalized. Full prompts and error taxonomies are provided in Appendix~\ref{sec:entailment_check_prompt} and Appendix~\ref{sec:decontextulization_check_prompt}.

We run this evaluation on FactScore-Bio, introduced in Section~\ref{sec:factuality_metric}, and LongFact~\citep{wei2024longform}. LongFact provides a harder setting for decomposition because it contains abstract concepts and complex objects across diverse domains such as business, psychology, and virology. We use the 200 GPT-4 responses from the LongFact subset curated by VeriScore. All texts are decomposed with the few-shot open information extraction pipeline in Section~\ref{sec:few_shot}, using the With Question / With Chunking mode and chunks of approximately 300 characters. We use Qwen3.5-397B-A17B as the decomposer. Table~\ref{tab:decomp_count} reports the number of extracted claim-level facts evaluated for each framework. Consistent with the claim-level variant in Section~\ref{sec:metric_evaluation}, a TriQua claim-level fact is one complete extracted fact: either a simple fact containing only a base triple, or a complex fact containing a base triple with attached qualifiers. Each claim-level fact is evaluated as one item in the entailment and decontextualization checks.
\begin{table}[t]
\centering
\scriptsize
\begin{tabular}{lccc}
\toprule
\textbf{Dataset} & \textbf{RefChecker} & \textbf{VeriScore} & \textbf{TriQua} \\
\midrule
FactScore-Bio & 3253 &3567 & 3565 (5622)  \\
LongFact      & 5198 & 10686 & 10032 (15216) \\
\bottomrule
\end{tabular}
\caption{
Number of extracted claim-level facts used for decomposition quality evaluation.
For TriQua, parentheses denote raw fact units before deduplication, counting base triples and qualifiers separately.
}
\label{tab:decomp_count}
\end{table}

\begin{table*}[!htbp] 
\centering
\footnotesize
\begin{tabular}{llcccc}
\toprule
\multirow{2}{*}{\textbf{Judge Model}} & \multirow{2}{*}{\textbf{Framework}} & \multicolumn{2}{c}{\textbf{Entailed (\%)}} & \multicolumn{2}{c}{\textbf{Decontextualized (\%)}} \\
\cmidrule(lr){3-4} \cmidrule(lr){5-6}
& & \textbf{FactScore-Bio} & \textbf{LongFact} & \textbf{FactScore-Bio} & \textbf{LongFact} \\
\midrule
\multirow{3}{*}{Qwen3.5-397B-A17B} 
& RefChecker & 99.48 & 92.86 & 98.40 & 91.29 \\
& VeriScore & \textbf{99.61} & 97.03 & \textbf{99.24} & \textbf{97.99} \\
& \textbf{TriQua (Ours)} & 99.58 & \textbf{98.54} & 97.84 & 95.39 \\
\midrule
\multirow{3}{*}{GLM-5-FP8 (744B)} 
& RefChecker & \textbf{99.23} & 90.63 & 97.69 & 92.36 \\
& VeriScore & 99.13 & 95.67 & \textbf{98.54} & \textbf{96.84} \\
& \textbf{TriQua (Ours)} & 99.13 & \textbf{98.05} & 97.34 & 94.15 \\
\bottomrule
\end{tabular}%
\caption{Decomposition quality evaluation comparing Entailment and Decontextualization across two datasets.}
\label{tab:decomp_quality}
\end{table*}

To reduce dependence on a single judge, we evaluate decomposition quality with two open-weight judge models from different families, Qwen3.5-397B-A17B and GLM-5-FP8 (744B)~\citep{glm5vibecodingagentic}. We interpret these judgments as comparative diagnostics rather than human gold labels. Table~\ref{tab:decomp_quality} shows that TriQua achieves consistently high entailment rates across datasets and judges. RefChecker performs well on the simpler FactScore-Bio setting but drops substantially on LongFact, suggesting that strict triples are less suitable for complex and abstract claims. VeriScore achieves the highest decontextualization rates, but its lower entailment rates indicate a different trade-off: manual inspection suggests that it sometimes over-contextualizes claims by adding information not grounded in the source text. Between the two judge models, GLM-5-FP8 is stricter in most cases. We report the Cohen's kappa~\citep{cohen_kappa} for the agreement ratio in Appendix~\ref{sec:cohen_kappa}.

\subsubsection{Qualifier usage evaluation}
\label{sec:qualifier_evaluation}
For the qualifier usage evaluation, we maintain the same experimental setup as the preceding analysis, utilizing the identical datasets, decomposer, and judge models. Table~\ref{tab:qualifier_stats} summarizes how often TriQua uses qualifiers with two datasets. 

\begin{table}[t]
\centering
\scriptsize
\begin{tabular}{lrrrrr}
\toprule
\textbf{Dataset} & \textbf{All} & \textbf{Simple} & \textbf{Complex} & \textbf{Qual.} & \textbf{Q./C} \\
\midrule
FactScore-Bio & 3565  & 1997 & 1568 & 2057 & 1.31 \\
LongFact      & 10032 & 6120 & 3912 & 5184 & 1.33 \\
\bottomrule
\end{tabular}
\caption{TriQua qualifier usage. All denotes claim-level facts. Q./C is the number of qualifiers per complex fact.}
\label{tab:qualifier_stats}
\end{table}

Following the usage rules established in Section~\ref{sec:qualifier_usage}, we classify qualifier misuse into four distinct error types: structural errors (where a qualifier improperly modifies another qualifier rather than the base triple), subject and object decomposition errors (over-bundling cases where a qualifier states a general, independent attribute about the subject or object instead of contextualizing the core relation), and semantic label errors (applying an incorrect label that alters the original meaning). The complete evaluation prompt is detailed in Appendix~\ref{sec:qualifier_misuse_check}. 

Table~\ref{tab:qualifier_misuse} reports binary qualifier-misuse rates, with detailed error-type distributions provided in Appendix~\ref{sec:error_rate}. Our qualifier-misuse judge is designed to detect harmful qualifier usage, not to enforce a single correct representation. A qualifier is marked as misuse when its attachment or label introduces ambiguity, changes the meaning of the target text, or makes subsequent verification unreliable. We therefore tolerate subject/object descriptive qualifiers that could also be represented as separate simple triples, as long as they remain locally tied to the base triple and do not introduce ambiguity. For example, from "Alberto Malesani managed Italian club Genoa." TriQua may extract \textit{(Alberto Malesani, managed, Genoa)-(type, Italian club)}. Although \textit{(Genoa, is, Italian club)} is also a valid simple fact, the qualifier is locally tied to the object \textit{Genoa} and preserves the intended meaning. Under this meaning-preserving criterion, qualifier misuse is rare across both datasets. On FactScore-Bio, semantic label error is the dominant error type for both judges. On LongFact, structural errors become more frequent, likely because longer and more abstract sentences in the chunks lead the decomposer to attach more qualifiers to a base triple. In some cases, a qualifier modifies another qualifier rather than the base triple itself.
\begin{table}[t]
\centering
\scriptsize
\begin{tabular}{lccc}
\toprule
\textbf{Dataset} & \textbf{Qwen Misuse (\%)} & \textbf{GLM Misuse (\%)} & \textbf{$\kappa$} \\
\midrule
FactScore-Bio & 1.34 & 1.08 & 0.41 \\
LongFact      & 2.22 & 1.61 & 0.42 \\
\bottomrule
\end{tabular}
\caption{
Binary qualifier-misuse rates and inter-judge agreement. $\kappa$ denotes Cohen's kappa and is computed over binary misuse labels from Qwen3.5-397B-A17B and GLM-5-FP8 as judge models.
}
\label{tab:qualifier_misuse}
\end{table}

\begin{table*}[!htbp]
\centering
\footnotesize 
\setlength{\tabcolsep}{4.5pt} 
\begin{tabular}{llccccc}
\toprule
\multirow{2}{*}{\textbf{Backbone Model}} & \multirow{2}{*}{\textbf{Framework}} & \multicolumn{4}{c}{\textbf{Subset F1 (\%)}} & \multirow{2}{*}{\textbf{Overall}} \\
\cmidrule(lr){3-6}
& & \textbf{AggreFact} & \textbf{SciFact} & \textbf{Cover} & \textbf{Hover} & \\
\midrule
\multirow{4}{*}{Qwen3.5-397B-A17B} 
& Few-Shot (No Decomp.) & 88.30 & 89.03 & 78.99 & 83.89 & 86.22 \\
\cmidrule{2-7}
& RefChecker & 82.89 & 85.39 & 73.9 & 77.65 & 81.11 \\
& VeriScore & 84.60 & \textbf{87.41} & 78.84 & \textbf{83.55} & 83.63 \\
& \textbf{TriQua (Ours)} & \textbf{87.98} & 87.23 & \textbf{80.02} & 82.79 & \textbf{85.91} \\
\midrule

\multirow{4}{*}{Qwen3.5-122B-A10B} 
& Few-Shot (No Decomp.) & 89.60 & 89.02 & 81.28 & 85.20 & 87.59 \\
\cmidrule{2-7}
& RefChecker & 83.91  & 85.98 & 72.59 & 78.59 & 81.49 \\
& VeriScore & 83.14 & \textbf{90.29} & 73.85 & \textbf{83.42} & 82.53 \\
& \textbf{TriQua (Ours)} & \textbf{85.56} & 85.37 & \textbf{77.94} & 79.58 & \textbf{83.46} \\
\bottomrule
\end{tabular}
\caption{F1 scores on the CLEARFACTS dataset across two backbone models. In all decomposition frameworks, the specified backbone model functions as both decomposer and verifier. TriQua is benchmarked against alternative decomposition pipelines and a direct-verification (no decomposition) baseline. Results are averaged across two independent runs for each setup. Full benchmark is presented in Appendix~\ref{sec:clearfacts_result_full}}.
\label{tab:clearfacts_results_part}
\end{table*}

\subsection{Evidence-based fact verification evaluation}
We evaluate short-form fact checking on CLEARFACTS~\citep{seo2025verifying}, which aggregates challenging datasets including AggreFact~\citep{tang-etal-2024-minicheck}, SciFact~\citep{wadden-etal-2020-fact}, CoverBench~\citep{jacovi2024coverbenchchallengingbenchmarkcomplex}, and Hover~\citep{jiang-etal-2020-hover}. Dataset statistics are provided in Appendix~\ref{sec:clearfacts_subsets_stats}. Since CLEARFACTS provides evidence for each claim, no retrieval is required. This setting is favorable to direct verification, as the verifier can judge the original claim against the given evidence directly. It does not test a potential advantage of decomposition for open-domain retrieval, where finer-grained subclaims may retrieve more targeted evidence. TriQua decomposes each claim using the Without Question / Without Chunking mode, verifies the resulting facts individually, and aggregates them into a final label.

Table~\ref{tab:clearfacts_results_part} reports results across two backbone models and dataset subsets. Full results are in Appendix~\ref{sec:clearfacts_result_full}. Among decomposition-based methods, TriQua generally outperforms RefChecker and VeriScore, especially on AggreFact and CoverBench. SciFact is less conclusive due to a small sample size of 89. VeriScore performs better on Hover, suggesting that sentences for fact representation offer a slight advantage in preserving the linguistic bridges and nuances required for multi-hop reasoning. Overall, TriQua's gains over RefChecker indicate that qualifiers help overcome limitations of purely triple-based fact representations for complex claims.

Few-shot direct verification without decomposition achieves the best overall performance, consistent with \citet{hu-etal-2025-decomposition}, who show that decomposition can hurt short-claim verification with strong verifiers due to cascading extraction errors. Appendix~\ref{sec:comparision_direct_triqua} shows that the main gap comes from lower recall on the Supported class: TriQua is more conservative because one refuted subclaim can make the whole claim Not Supported. In the Qwen3.5-122B-A10B setting, where this gap is largest, our manual analysis of 50 error cases shows that the main sources are flawed or ambiguous original claim annotations and TriQua over-strictness (Appendix~\ref{sec:triqua_qualitative_error_analysis}).

\section{Conclusion}
We introduced TriQua, a framework designed to resolve the fundamental trade-off between atomicity and context preservation in the "decompose-then-verify" paradigm for LLM factuality evaluation. While previous methods struggle to balance fine-grained analysis with necessary contextual modifiers, TriQua adaptively models facts based on their complexity. Simple claims are extracted as standard atomic triples, whereas complex claims are represented as hyperrelational facts that pair a base triple with a set of independent contextual qualifiers, such as temporal or spatial constraints. This highly structured approach ensures that essential context is retained for accurate information retrieval and verification without violating the principle of atomicity. Furthermore, TriQua's clean separation of base triples and qualifiers allows its verification process to explicitly pinpoint exactly which component of a claim is unsupported. This prevents the unfair penalization of an entire complex statement due to a single faulty modifier, thereby providing granular explainability for error detection. Alongside this framework, we propose TriQuaScore, a relation-based precision metric that quantifies the factuality of these distinct fact units. Empirical evaluations confirm the system's effectiveness: TriQua achieves robust decomposition quality by maintaining high entailment and reliable decontextualization. It outperforms existing decomposed-based baselines such as RefChecker and VeriScore, making it a highly capable, context-aware tool for systematic factuality evaluation.

\section*{Limitations}
Our factuality metric evaluation is mainly conducted on FactScore-Bio, which is entity-centric and may favor structured decomposition frameworks such as TriQua. Extending the evaluation to less entity-centric long-form generations remains challenging because few datasets provide fine-grained human factuality annotations at the atomic-fact level, and creating such annotations is costly.

Our decomposition-quality analysis also relies primarily on LLM-as-a-judge evaluation. Although we use two judge models from different families and report inter-judge agreement, moderate Cohen's $\kappa$ values indicate that automatic judgments, especially fine-grained error categories, remain noisy.

TriQua's fine-grained component-level verification can improve error localization, but it can also be too strict with checking each component of base triple and qualifiers, which may increase false positives. In addition, our current evidence retrieval setting is mostly closed-domain or based on provided evidence. The potential benefit of qualified decomposition for open-domain retrieval, where fine-grained queries may retrieve more targeted evidence, requires further study.
\section*{Ethics Statement}
During the preparation of this paper, the authors utilized Gemini Pro and GPT-5.5 to correct grammatical errors and refine the fluency of the text. Additionally, large language models were employed to iteratively polish the instructions within our prompts.


\bibliography{custom}

\appendix
\onecolumn
\section{Appendix}

\subsection{Decomposition modes.}
\label{sec:decomposition_modes}
TriQua supports four prompt variants for decomposition depending on whether a user question is available and document chunking is necessary. Table~\ref{tab:decomposition_modes} shows a detailed description of each mode.
\begin{table}[!htbp]
\renewcommand{\arraystretch}{1.4} 
\small
\renewcommand{\tabularxcolumn}[1]{>{\raggedright\arraybackslash}m{#1}}

\begin{tabularx}{\linewidth}{>{\hsize=0.72\hsize\bfseries}X | >{\hsize=1.28\hsize}X}
\hline
Mode & \textbf{Description} \\
\hline
With Question / With Chunking & For decomposing document-level texts generated in response to a specific question, utilizing both the question and the preceding chunk for context. \\
\hline
Without Question / With Chunking & For decomposing long-form documents without a guiding question (e.g., general summaries), where contextualization relies solely on the preceding text chunk. \\
\hline
With Question / Without Chunking & For decomposing short, sentence- or paragraph-level texts where the question provides necessary context, but text chunking is not required. \\
\hline
Without Question / Without Chunking & For decomposing standalone short texts into facts without the question or chunking. \\
\hline
\end{tabularx}
\caption{Four decomposition modes used by TriQua. ``Question'' denotes any task-level input conditioning the generation, including a user question, instruction, or prompt.}
\label{tab:decomposition_modes}
\end{table}
\subsection{Prompts}
\subsubsection{Prompt for decomposition mode "With Question / With Chunking"}
\label{sec:decomp_promp_wq_wc}
\begin{tcblisting}{
  colback=gray!5,
  colframe=gray!50!black,
  listing only,
  breakable,
  size=fbox, % Reduces gray padding
  title=Decomposition Prompt: With Question With Chunking,
  title after break={Decomposition Prompt: With Question With Chunking (Continued)},
  listing options={
    %basicstyle=\ttfamily\fontsize{8pt}{8.5pt}\selectfont,
    basicstyle=\ttfamily\scriptsize\linespread{0.8}\selectfont,
    breaklines=true,
    breakindent=0pt,        % Forces wrapped line indent to 0
    breakautoindent=false,  % Turns off the code-continuation indent
    columns=fullflexible,   % CRUCIAL: Fixes weird word spacing and alignment gaps!
    upquote=true            % Brings back the missing double quotes
  }
}
# Task: Decompose Target Text for Factuality Evaluation
Your task is to decompose the **Target Text** into atomic facts for factuality evaluation. Follow these steps:

## Steps to Follow:
### 1. Establish Context:
- Use the accompanying question and context to establish context for the **Target Text**.
- **Do NOT decompose the question and context**.
    - **Exception**: You may use the **Question** and **Context** to form a fact only if the **Target Text** is a short phrase that cannot form a fact on its own (e.g., a name, location). In this case you may synthesize a fact using the **Question**, **Context** and **Target Text**.

### 2. Extract Objective Sentences:
- Goal: Identify objective sentences (or parts of sentences) that make specific claims about the world which could theoretically be verified as true or false by an external fact-checker. You are not asked to determine whether the claims are true or false, only to extract them.
- Inclusion Criteria:
    - Empirical Reality: Physical states, events, data, statistics, scientific phenomena, definitions, or laws (e.g., "Water boils at 100 degrees").
    - Attributed Stance: If the text reports an opinion, treat the existence of the opinion as the fact (e.g., "Critics described the movie as thrilling" --> Fact: "Critics expressed a positive view").
    - The "Buried Fact" Rule: If a sentence mixes opinion and fact (e.g., "The innovative Company X released Product Y"), extract only the factual core ("Company X released Product Y").
    - General Verifiability (Catch-All): Any other claim that implies a specific, checkable state of the world.
- Exclusion Criteria:
    - Opinions: Subjective judgements without attribution (e.g., "This is the best approach").
    - Subjective & Speculative: Pure opinion, value judgements, predictions, or hypothetical scenarios (e.g., "I believe this will work").
    - Pure Advice/Imperatives: Recommendations (e.g., "You should verify this").
    - Meta-Discourse: Navigational text (e.g., "Here is the summary," "As an AI...").
    - Negative Capability: Statements about what the model cannot find.


### 3. Decompose into Atomic Facts (Open-IE Style):
- Break down the objective sentences into fundamental atomic facts using one of the following formats:
   - **Relation Triplet (Simple Facts)**:
      ```json
      {"base_triple": ["subject", "relation", "object"]}
      ```
   - **Hyper-Relational Tuple (Base Triplet + Qualifiers)**:
     Use this when additional details (e.g., context, causality, sequence, provenance) are needed to clarify the base triple.
      ```json
      {"base_triple": ["subject", "relation", "object"],
       "qualifiers": [{"qualifier-label-1": "qualifier-value-1"}, {"qualifier-label-2": "qualifier-value-2"}]}
      ```
- Open-IE conventions for relations and qualifier labels:
    - Relation (in base_triple): copy the core verb or relational phrase directly from the text, trimming only what is necessary. Keep modal or hedge words ("can", "may" etc.) and continuous/aspectual verbs (e.g., "continues to," "began to") intact within the relation. Do not strip grammatical markers or tense to create qualifiers.
    - Embed Negations in the Relation: Include negation words (e.g., "not," "cannot," "no") directly inside the base relation. Do NOT extract a positive relation and push the negation into a qualifier
    - Qualifier label: a concise, self-explanatory role word or short phrase. Labels need not match any fixed ontology but should clearly convey how the value contextualize the base triple.
    - Qualifier value: the corresponding phrase, date, number, entity or clause extracted from the text.


## Guidelines:
### 1. Accuracy and Completeness:
- Ensure each fact is standalone and complete. Do not add information not explicitly stated in the response.
- **Resolve coreference** (e.g., pronouns, abbreviations) by using full names or complete identifiers rather than generic definite noun phrases (e.g., "the scientist", "the teacher") to avoid ambiguity. **Do not use pronouns**.
- **Disambiguate Definite Phrases**: If a definite noun phrase must be used (e.g., "the chemical plant," "the president," "the fire"), add necessary modifiers from the text (e.g., a prepositional phrase or embedded clause) to ensure the entity is unambiguous (e.g., "The chemical plant in China's Fujian province", "President Bush", "The fire at the chemical plant in China's Fujian province").
- **Resolve Missing Baselines for Relative Terms**: If a sentence contains relative terms---including standard comparatives (higher, more), state-changes (improved, reduced), additives (additional, extra), or temporal relatives (newer, previous), but omits the baseline, you MUST look at the Question and Context to find the baseline entity. If the baseline is available, add it using a qualifier like `compared to`, `in addition to` etc.
- Do not expand abbreviations or names unless explicitly stated in the input. Stick strictly to the explicit facts provided.
### 2. Explicit Information Only:
- Extract only explicitly stated information. Do not introduce implicit facts, assumptions, or interpretations.
### 3. Qualifiers Categories:
- **Validity Context**: Define the validity of the base triplet based on time, location, or specific conditions.
- **Causality**: Indicate the cause-and-effect or the reason for the base triplet.
- **Sequence**: Describe the order or sequence of events (e.g., replaces, replaced by).
- **Provenance**: Provide information about how a fact was determined (e.g., source or method).
- **Annotation**: Add additional information to the statement, including constraint annotations and other miscellaneous attributes that provide extra context.
### 4. Qualifiers Usage:
- Prioritize qualifiers to decompose complex statements, especially when dealing with time, location, method, scope, conditions, or reasons.
- A base triplet can have multiple qualifiers, which form the set `qualifiers`.
- Qualifiers should directly modify or clarify the base triplet and not other qualifiers.
- Do not use qualifiers to represent separate facts that are not attributes of the base triplet.
### 5. Atomicity (Separable vs. Inseparable):
- Break compound subjects, objects, or qualifier values connected by "and," "or," or commas into separate facts ONLY IF each resulting fact is independently verifiable and represents a distinct claim.
- Keep Inseparable Conjunctions Intact: Do NOT split elements if the division alters the core meaning or creates unverifiable fragments.
### 6. Decomposing Complex Sentences and Clauses
Use this section to decide whether a clause becomes a new fact or a qualifier.
- Attribution (e.g., "said," "found," "estimates"):
    - Rule: Treat the attribution as the main fact. The source is the subject, the attribution verb (e.g., "estimates") is the relation, and the reported claim is the object. This adheres to Step 2 (Objective Sentences), as the act of reporting is the objective fact.
- Contextual vs. Descriptive Clauses: To handle all other clauses (relative, appositive, subordinate):
    - Becomes a QUALIFIER (Direct Context): Rule: If the clause provides direct context for the base triple's assertion (e.g., when, where, why, how, or under what condition the S-R-O relationship holds), use it as a qualifier.
    - Becomes a NEW FACT (General Property): Rule: If the clause states a general property or a separate fact about the `subject` or `object` (rather than contextualizing the main `relation`), decompose it into a new, independent `base_triple`.
### 7. Avoid Redundancy:
- Do not repeat the same fact with different phrasing or by reversing subject and object.
### 8. Conciseness:
- Do not include introductory phrases, explanations, or additional notes. Only generate the decomposed facts in the required format for the **Target Text**.
- Use qualifiers sparingly and only when necessary to refine the base triplet.

## Examples:
### Example 1:
#### Question:
Discuss the role of Aluminum Corporation of China Limited in the global aluminium industry.
#### Context:
The aluminium industry is a cornerstone of modern manufacturing. One of its most significant players is Aluminum Corporation of China Limited.

#### **Target Text**:
"""
Aluminum Corporation of China Limited (known as Chalco; parent group Aluminum Corporation of China is known as Chinalco) (SEHK: 2600, NYSE: ACH, SSE:601600), is a state-owned, multinational aluminium company headquartered in Beijing, People's Republic of China. It holds a near monopoly on alumina refining in the domestic market.
"""
#### Decomposed Facts:
```json
[
  {"base_triple": ["Aluminum Corporation of China Limited", "known as", "Chalco"]},
  {"base_triple": ["Aluminum Corporation of China Limited", "parent organization", "Aluminum Corporation of China"]},
  {"base_triple": ["Aluminum Corporation of China", "known as", "Chinalco"]},
  {"base_triple": ["Aluminum Corporation of China Limited", "listed on", "SEHK"], "qualifiers": [{"stock code": "2600"}]},
  {"base_triple": ["Aluminum Corporation of China Limited", "listed on", "NYSE"], "qualifiers": [{"stock code": "ACH"}]},
  {"base_triple": ["Aluminum Corporation of China Limited", "listed on", "SSE"], "qualifiers": [{"stock code": "601600"}]},
  {"base_triple": ["Aluminum Corporation of China Limited", "type", "state-owned company"]},
  {"base_triple": ["Aluminum Corporation of China Limited", "type", "multinational company"]},
  {"base_triple": ["Aluminum Corporation of China Limited", "operates in", "aluminium industry"]},
  {"base_triple": ["Aluminum Corporation of China Limited", "headquartered in", "Beijing"]},
  {"base_triple": ["Beijing", "located in", "People's Republic of China"]},
  {"base_triple": ["Aluminum Corporation of China Limited", "holds a near monopoly on", "alumina refining"],
   "qualifiers": [{"applies to": "the domestic market"}]}
]
```

### Example 2:
#### Question:
What are the main focuses of Attachment Theory?
#### Context:
Attachment Theory, originally developed by John Bowlby and later expanded upon by Mary Ainsworth, is a psychological, evolutionary, and ethological theory concerning relationships between humans.

#### **Target Text**:
"""
The theory primarily focuses on the nature of the close emotional bond that develops between infants and their primary caregivers and how this bond impacts the individual's emotional development, social relationships, and psychological well-being throughout their life.
"""
#### Decomposed Facts:
```json
[
  {"base_triple": ["Attachment Theory", "primarily focuses on", "the nature of the close emotional bond that develops between infants and their primary caregivers"]},
  {"base_triple": ["Attachment Theory", "primarily focuses on", "how the close emotional bond that develops between infants and their primary caregivers impacts the individual's emotional development throughout their life"]},
  {"base_triple": ["Attachment Theory", "primarily focuses on", "how the close emotional bond that develops between infants and their primary caregivers impacts the individual's social relationships throughout their life"]},
  {"base_triple": ["Attachment Theory", "primarily focuses on", "how the close emotional bond that develops between infants and their primary caregivers impacts the individual's psychological well-being throughout their life"]}
]
```

### Example 3:
#### Question:
What is the maximum range of the Boeing 777?
#### Context:
No extra context information provided.

#### **Target Text**:
"""
I cannot provide a single answer to this question because the maximum range depends on the specific variant of the aircraft. Today, these aircraft, particularly the extended-range models, operate worldwide. For instance, the Boeing 777-200ER has a maximum range of 7,065 nautical miles. Following a major engine upgrade, the 777-300ER model achieves a range of approximately 7,370 nautical miles. Furthermore, on November 10, 2005, a 777-200LR set a record by flying 11,664 nautical miles from Hong Kong to London. Please specify which model you are interested in so I can assist you further.
"""
#### Decomposed Facts:
```json
[
  {"base_triple": ["Boeing 777 aircraft", "operates", "worldwide"],
   "qualifiers": [{"applies particularly to": "extended-range models"}]},
  {"base_triple": ["Boeing 777-200ER", "has maximum range", "7,065 nautical miles"]},
  {"base_triple": ["Boeing 777-300ER model", "achieves a range of", "approximately 7,370 nautical miles"],
   "qualifiers": [{"after": "a major engine upgrade"}]},
  {"base_triple": ["Boeing 777-200LR", "set a record by flying", "11,664 nautical miles"],
   "qualifiers": [{"date": "November 10, 2005"}, {"departure location": "Hong Kong"}, {"destination location": "London"}]}
]
```

### Example 4:
#### Question:
Do you think the new urban development plan for the city center will be successful?
#### Context:
The city council recently proposed a controversial plan to redesign the downtown district to prioritize pedestrian traffic over vehicles.

#### **Target Text**:
"""
In my opinion, the new plan is a recipe for disaster that will likely destroy local businesses. It feels like the city council is ignoring the needs of commuters, and I suspect this initiative is just a way to push a specific political agenda.
"""
#### Decomposed Facts:
```json
[]
```

### Example 5:
#### Question:
Why is it important to understand function properties in mathematics, and how do convex functions play a key role in this understanding?
#### Context:
Understanding the properties of functions is fundamental in mathematics, particularly when analyzing their behavior and applications. One important class of functions is convex functions, which have unique geometric characteristics.

#### **Target Text**:
"""
A real-valued function is called convex if the line segment between any two distinct points on the graph of the function lies above or on the graph between the two points. Data scientists frequently optimize these functions using gradient descent algorithms, except in discrete state spaces.
"""
#### Decomposed Facts:
```json
[
  {"base_triple": ["A real-valued function", "is called", "convex"],
   "qualifiers": [{"condition": "if the line segment between any two distinct points on the graph of the function lies above or on the graph between the two points"}]},
  {"base_triple": ["Data scientists", "frequently optimize", "convex functions"],
   "qualifiers": [{"method": "using gradient descent algorithms"}, {"exception": "in discrete state spaces"}]}
]
```

### Example 6:
#### Question:
Tell me about the recent chemical industrial accident in China's Fujian province.
#### Context:
Authorities are investigating a recent industrial accident in China's Fujian province.

#### **Target Text**:
"""
An explosion at a chemical plant in China's Fujian province injured six people and caused a large fire. The plant produces a potentially carcinogenic chemical called paraxylene. The explosion occurred at an oil storage facility due to an oil leak, but no toxic chemical spill has been reported. The fire has been brought under control, and those injured have been sent to the hospital.
"""
#### Decomposed Facts:
```json
[
  {"base_triple": ["Fujian province", "located in", "China"]},
  {"base_triple": ["The explosion at the chemical plant in China's Fujian province", "injured", "six people"]},
  {"base_triple": ["The explosion at the chemical plant in China's Fujian province", "caused", "a large fire"]},
  {"base_triple": ["The chemical plant in China's Fujian province", "produces", "paraxylene"]},
  {"base_triple": ["Paraxylene", "is", "a chemical"]},
  {"base_triple": ["Paraxylene", "is", "potentially carcinogenic"]},
  {"base_triple": ["The explosion at the chemical plant in China's Fujian province", "occurred at", "an oil storage facility"],
   "qualifiers": [{"cause": "oil leak"}]},
  {"base_triple": ["A toxic chemical spill from the explosion at the chemical plant in China's Fujian province", "has not been", "reported"]},
  {"base_triple": ["The fire at the chemical plant in China's Fujian province", "was", "brought under control"]},
  {"base_triple": ["The six people injured in the explosion at a chemical plant in China's Fujian province", "were taken to", "a hospital"]}
]
```

### Example 7:
#### Question:
Why are scientific discoveries important for biology and medicine, and how are they recognized?
#### Context:
Scientific breakthroughs often revolutionize our understanding of biology and medicine. One such groundbreaking discovery was the identification of RNA interference (RNAi), a mechanism that has transformed genetic research and therapeutic development.

#### **Target Text**:
"""
Andrew Fire was awarded the 2006 Nobel Prize for Physiology or Medicine, along with Craig C. Mello, for the discovery of RNA interference (RNAi).
"""
#### Decomposed Facts:
```json
[
  {"base_triple": ["Andrew Fire", "award received", "Nobel Prize for Physiology or Medicine"],
   "qualifiers": [{"time": "2006"}, {"shared with": "Craig C. Mello"}, {"reason": "discovery of RNA interference (RNAi)"}]}
]
```

### Example 8:
#### Question:
What level of availability does Google Cloud guarantee, and how do they achieve it?
#### Context:
Google Cloud operates a worldwide network of redundant, geographically distributed data centers designed to maximize reliability.
#### **Target Text**:
"""
The cloud provider's Service Level Agreement (SLA) guarantees 99.99% uptime, including scheduled maintenance periods, by utilizing redundant geographically-dispersed data centers. This high availability prevents service interruptions, ensuring continuous business operations for global clients. Furthermore, to maintain strict data boundaries and comply with international privacy regulations, no access to the physical servers is possible for external contractors.
"""
#### Decomposed Facts:
```json
[
  {"base_triple": ["Google Cloud's Service Level Agreement (SLA)", "guarantees", "99.99% uptime"],
   "qualifiers": [{"including": "scheduled maintenance periods"}, {"mechanism": "by utilizing redundant geographically-dispersed data centers"}]},
  {"base_triple": ["Google Cloud's 99.99% uptime", "prevents", "service interruptions"],
   "qualifiers": [{"result": "continuous business operations for global clients"}]},
  {"base_triple": ["External contractors", "have no possible access to", "Google Cloud's physical servers"],
   "qualifiers": [{"reason": "to maintain strict data boundaries"}, {"reason": "to comply with international privacy regulations"}]}
]
```

### Example 9:
#### Question:
Compare recent textbooks in reinforcement learning to the book "Reinforcement Learning: An Introduction".
#### Context:
Textbooks in reinforcement learning vary in depth, ranging from introductory guides focused on conceptions to mathematically rigorous books designed for advanced theories.
#### **Target Text**:
"""
The book "Reinforcement Learning: An Introduction" was jointly authored by Richard S. Sutton and Andrew G. Barto. This book can provide a foundational understanding of MDPs, TD learning, and other core concepts. However, Szepesvari's book delivers tighter theoretical bounds. "Deep Reinforcement Learning Hands-On" includes more code examples.
"""
#### Decomposed Facts:
```json
[
  {"base_triple": ["The book 'Reinforcement Learning: An Introduction'", "was authored by", "Richard S. Sutton"]},
  {"base_triple": ["The book 'Reinforcement Learning: An Introduction'", "was authored by", "Andrew G. Barto"]},
  {"base_triple": ["The book 'Reinforcement Learning: An Introduction' by Richard S. Sutton and Andrew G. Barto", "can provide a foundational understanding of", "core concepts"],
   "qualifiers": [{"including": "MDPs"}, {"including": "TD learning"}]},
  {"base_triple": ["The book by Szepesvari", "delivers", "tighter theoretical bounds"],
   "qualifiers": [{"compared to": "the book 'Reinforcement Learning: An Introduction' by Richard S. Sutton and Andrew G. Barto"}]},
  {"base_triple": ["The book 'Deep Reinforcement Learning Hands-On'", "includes", "more code examples"],
   "qualifiers": [{"compared to": "the book 'Reinforcement Learning: An Introduction' by Richard S. Sutton and Andrew G. Barto"}]}
]
```

### Example 10:
#### Question:
What are some notable works published by British author J.K. Rowling?
#### Context:
J.K. Rowling is a British author and philanthropist.
#### Target Text:
"""
She wrote the fantasy novel "Harry Potter", the crime thriller "The Cuckoo's Calling", and the stage play "Harry Potter and the Cursed Child".
"""
#### Decomposed Facts:
```json
[
  {"base_triple": ["J.K. Rowling", "wrote", "'Harry Potter'"]},
  {"base_triple": ["'Harry Potter' by J.K. Rowling", "is", "a fantasy novel"]},
  {"base_triple": ["J.K. Rowling", "wrote", "'The Cuckoo's Calling'"]},
  {"base_triple": ["'The Cuckoo's Calling' by J.K. Rowling", "is", "a crime thriller"]},
  {"base_triple": ["J.K. Rowling", "wrote", "'Harry Potter and the Cursed Child'"]},
  {"base_triple": ["'Harry Potter and the Cursed Child' by J.K. Rowling", "is", "a stage play"]}
]
```


## Now Your Turn:
#### Question:
$question
#### Context:
$context

#### **Target Text**:
"""
$target_text
"""
#### Decomposed Facts:

\end{tcblisting}

\subsubsection{Prompt for decomposition mode "Without Question / Without Chunking"}
To avoid redundancy, we omit the full text of the ``Without Question / Without Chunking'' prompt, as the core instructions are largely identical to the default prompt detailed in Appendix \ref{sec:decomp_promp_wq_wc}. 

The primary modification is that the model receives only the \textbf{Target Text} as input, making it the sole source of information for decontextualization. Consequently, the input fields in the few-shot examples are also adjusted to remove the Question and Context fields. We illustrate this structural adaptation below using a modified version of Example 8 from the original prompt:
\label{sec:decomp_promp_woq_woc}
\begin{tcblisting}{
  colback=gray!5,
  colframe=gray!50!black,
  listing only,
  breakable,
  size=fbox, % Reduces gray padding
  title=Decomposition Prompt: Without Question Without Chunking,
  title after break={Decomposition Prompt: Without Question Without Chunking (Continued)},
  listing options={
    %basicstyle=\ttfamily\fontsize{8pt}{8.5pt}\selectfont,
    basicstyle=\ttfamily\scriptsize\linespread{0.8}\selectfont,
    breaklines=true,
    breakindent=0pt,        % Forces wrapped line indent to 0
    breakautoindent=false,  % Turns off the code-continuation indent
    columns=fullflexible,   % CRUCIAL: Fixes weird word spacing and alignment gaps!
    upquote=true            % Brings back the missing double quotes
  }
}
### Example 8:
#### Target Text:
Google Cloud's Service Level Agreement (SLA) guarantees 99.99% uptime, including scheduled maintenance periods, by utilizing redundant geographically-dispersed data centers. This high availability prevents service interruptions, ensuring continuous business operations for global clients. Furthermore, to maintain strict data boundaries and comply with international privacy regulations, no access to the physical servers is possible for external contractors.
#### Decomposed Facts:
```json
[
  {"base_triple": ["Google Cloud's Service Level Agreement (SLA)", "guarantees", "99.99% uptime"],
   "qualifiers": [{"including": "scheduled maintenance periods"}, {"mechanism": "by utilizing redundant geographically-dispersed data centers"}]},
  {"base_triple": ["Google Cloud's 99.99% uptime", "prevents", "service interruptions"],
   "qualifiers": [{"result": "continuous business operations for global clients"}]},
  {"base_triple": ["External contractors", "have no possible access to", "Google Cloud's physical servers"],
   "qualifiers": [{"reason": "to maintain strict data boundaries"}, {"reason": "to comply with international privacy regulations"}]}
]
```
\end{tcblisting}

\subsubsection{Prompt for triqua verification and annotation}
\label{sec:verification_annotation_prompt}
\begin{tcblisting}{
  colback=gray!5,
  colframe=gray!50!black,
  listing only,
  breakable,
  size=fbox, % Reduces gray padding
  title=Prompt for Triqua Verification and Annotation,
  title after break={Prompt for Triqua Verification and Annotation (Continued)},
  listing options={
    %basicstyle=\ttfamily\fontsize{8pt}{8.5pt}\selectfont,
    basicstyle=\ttfamily\scriptsize\linespread{0.8}\selectfont,
    breaklines=true,
    breakindent=0pt,        % Forces wrapped line indent to 0
    breakautoindent=false,  % Turns off the code-continuation indent
    columns=fullflexible,   % CRUCIAL: Fixes weird word spacing and alignment gaps!
    upquote=true            % Brings back the missing double quotes
  }
}
# System Role & Objective
You are an expert Fact Verification System. Your task is to perform a step-by-step verification of a **Structured Fact** against a provided set of **Evidence** text snippets. You must reason through the relationship, assign a verification label, and identifying specific error spans if the fact is not supported.

# Input Data
You will receive:
1. Evidence: A collection of text snippets acting as the sole source of truth.
2. Structured Fact To Verify: This will appear in one of the two formats:
   - A simple triplet: (subject, predicate, object)
   - A triplet with qualifiers: A primary triplet `(subject, predicate, object)` appended with hyphen-separated qualifier tuples `-(label, value)-...-(label, value)`. The qualifiers provide additional context to the main triplet.


# Verification Process
Follow this exact Chain-of-Thought process:

## Step 1: Analyze & Reason
- Compare each component of the `Structured Fact To Verify`, `(Subject, Predicate, Object)-(label,value)...`, against the Evidence.
- Determine if the Evidence supports, contradicts, or fails to mention the information in the fact.
- Formulate a brief, concise summary about this comparison. This will be your `reasoning_process`.
- **Crucial**: Do not stop if you find one error. Continue checking the remaining components to see if they are also incorrect or missing.
- **Note on Reasoning**: You are permitted and expected to use reasoning, inference, and external common sense to understand the evidence, fill in minor gaps, and verify the fact. Focus on the core assertion, not minor discrepancies like synonyms or phrasing that don't change the fact's meaning.

## Step 2: Determine Label
Based on Step 1, assign one of the following labels:
- Entailment: The `Evidence` clearly and fully supports the `Structured Fact To Verify`.
- Contradiction: At least one component of the `Structured Fact To Verify` is directly contradicted by the `Evidence`.
- Neutral: None of the components of the `Structured Fact To Verify` are contradicted by the `Evidence`. At least one component is missing from or unsubstantiated by the evidence.

## Step 3: Conditional Error Annotation
- IF Entailment: Do nothing further.
- IF Contradiction or Neutral: you must list **ALL components from Structured Fact To Verify** that failed verification in the `error_span_annotation` list.

# Output Format
Output a single JSON object.
Scenario A: Entailment
```json
{
    "reasoning_process": "Your brief reasoning process explaining the entailment.",
    "verification_label": "Entailment"
}
```

Scenario B: Contradiction or Neutral
```json
{
    "reasoning_process": "Your brief reasoning process explaining the contradiction or lack of evidence.",
    "verification_label": "Contradiction OR Neutral",
    "error_span_annotation": ["Identified span 1 from Structured Fact", "Identified span 2 from Structured Fact", "..."]
}
```

# Your Turn
## Evidence:
$evidence
## Structured Fact To Verify:
$claim
## Output:
\end{tcblisting}

\subsubsection{Prompt for entailment check}
\label{sec:entailment_check_prompt}
\begin{tcblisting}{
  colback=gray!5,
  colframe=gray!50!black,
  listing only,
  breakable,
  size=fbox, % Reduces gray padding
  title=Prompt for Entailment Check,
  title after break={Prompt for Entailment Check (Continued)},
  listing options={
    %basicstyle=\ttfamily\fontsize{8pt}{8.5pt}\selectfont,
    basicstyle=\ttfamily\scriptsize\linespread{0.8}\selectfont,
    breaklines=true,
    breakindent=0pt,        % Forces wrapped line indent to 0
    breakautoindent=false,  % Turns off the code-continuation indent
    columns=fullflexible,   % CRUCIAL: Fixes weird word spacing and alignment gaps!
    upquote=true            % Brings back the missing double quotes
  }
}
# Task
You are a strict Logic & Factuality Judge. Your task is to verify if a "Decomposed Fact" (a single extracted fact or claim) is logically supported by the entirety of the "Source Text", utilizing the "Context" if provided.

## Critical Distinctions
- Evaluate Globally: You must evaluate the Decomposed Fact against the entire Target Text. If the text contains conditions, hedges, or scope limits anywhere that invalidate the absolute truth of the Decomposed Fact, you must fail it.
- Ignore Reference Ambiguity: Do not penalize pronouns ("He", "It") if the logic holds true assuming the reference is correct.
- Enforce Modality & Conditions: You must mark as Non-Entailment if the decomposed fact drops logical conditions ("If", "Unless") or modality ("Might", "Could"), as this changes hypothetical statements into absolute facts.
- Allow Circumstantial Dropping: You must mark as Entailment decomposed facts that drop non-essential details (Dates, Locations, Instruments), provided the core action or relationship remains unchanged.

## Inputs
- `Context`: Background information (e.g., a preceding question, a prompt, or surrounding metadata). This may be empty. If provided, use it to resolve references or understand what the Target Text is addressing.
- `Source Text`:  The full source text (e.g., an answer, an article, a generated response) from which the fact was extracted.
- `Decomposed Fact`: The individual claim/fact (can be a sentence, triple, or triple with qualifiers) that needs to be evaluated. In case of triple with qualifiers, consider the entire fact including qualifiers, which contextualizes the triple.



## Verification Guidelines

### 1. The Decontextualization Protocol
Distinguish between Valid Decontextualization (Enrichment) and Inventing Facts (Hallucination).

A. Valid Decontextualization (Enrichment) -> Entailment
- Rule: The Decomposed Fact adds information available from Context or Source Text to resolve pronouns.
- Context: "What had Marie Curie discovered?" | Source Text: "She discovered radium."
- Decomposed Fact: "Marie Curie discovered radium." -> Entailment

B. Hallucination (Invalid) -> Non-Entailment (error_category: **HALLUCINATION**)
- Rule: The Decomposed Fact contains entities, names, or other information found in neither the Source Text nor the Context.
- Context: "What has Marie Curie discovered?" | Source Text: "She discovered radium."
- Decomposed Fact: "Marie Curie discovered radium in 1898." -> Non-Entailment

### 2. The List Decomposition Rule (Description vs. Definition)
A. Description Mode -> Entailment
- Rule: Source Text describes features or examples using non-exclusive verbs (e.g., *include, feature, have, use*). Extracting a single item from the list is valid.
    - Source Text: "The new car model features built-in GPS and an anti-lock braking system."
    - Decomposed Fact: "The new car model features built-in GPS." -> Entailment

B. Definition Mode -> Non-Entailment (error_category: **FALSE_EXCLUSIVITY**)
- Rule: Text defines the complete identity or exhaustive composition of a subject using exclusive verbs (e.g., *consist of, comprise, is composed of*). Extracting only a partial list implies the extracted part equals the entire whole.
    - Source Text: "A standard water molecule is composed of two hydrogen atoms and one oxygen atom."
    - Decomposed Fact: "A standard water molecule is composed of two hydrogen atoms." -> Non-Entailment

### 3. Dropped Conditions & Modality -> Non-Entailment (error_category: **CONDITION_ERROR**)
- Modality: Dropping might, could, probably, likely is a failure.
    - Source Text: "The software update could potentially fix the bug."
    - Decomposed Fact: "The software update fixes the bug." -> Non-Entailment
- Conditions: Dropping if, unless, assuming is a failure.
    - Source Text: "Customers will receive a full refund if they cancel their subscription within one week."
    - Decomposed Fact: "Customers will receive a full refund." -> Non-Entailment
- Exception (Circumstantial Modifiers): Dropping non-essential descriptive details (dates, locations, etc.) is acceptable and should be marked as Entailment, as long as the core assertion doesn't change.
    - Source Text: "Alice played for Team B in 2008."
    - Decomposed Fact: "Alice played for Team B." -> Entailment

### 4. Over-Generalization -> Non-Entailment (error_category: **PRECISION_ERROR**)
- Quantifiers: Dropping words like some, many, frequently, mostly to create a universal claim.
    - Source Text: "Most restaurants are closed on Sundays."
    - Decomposed Fact: "Restaurants are closed on Sundays." -> Non-Entailment
- Critical Modifiers: Dropping adjectives that restrict the subject to a specific subset (e.g., "Electric cars" -> "Cars").
    - Source Text: "Public universities in Germany offer tuition-free education."
    - Decomposed Fact: "Universities in Germany offer tuition-free education." -> Non-Entailment


## Output Format

Case 1: Entailment
```json
{
  "reasoning_process": "Your brief reasoning process explaining the entailment.",
  "entailment_label": "Entailment"
}
```

Case 2: Non-Entailment
```json
{
  "reasoning_process": "Your brief reasoning process explaining the non-entailment",
  "entailment_label": "Non-Entailment",
  "error_category": "HALLUCINATION | FALSE_EXCLUSIVITY | CONDITION_ERROR | PRECISION_ERROR"
}
```

### Input Data
#### Context
$context

#### Source Text
$source_text

#### Decomposed Fact
$decomposed_fact

### Output
\end{tcblisting}

\subsubsection{Prompt for decontextualization check}
\label{sec:decontextulization_check_prompt}
\begin{tcblisting}{
  colback=gray!5,
  colframe=gray!50!black,
  listing only,
  breakable,
  size=fbox, % Reduces gray padding
  title=Prompt for Decontextualization Check,
  title after break={Prompt for Decontextualization Check (Continued)},
  listing options={
    %basicstyle=\ttfamily\fontsize{8pt}{8.5pt}\selectfont,
    basicstyle=\ttfamily\scriptsize\linespread{0.8}\selectfont,
    breaklines=true,
    breakindent=0pt,        % Forces wrapped line indent to 0
    breakautoindent=false,  % Turns off the code-continuation indent
    columns=fullflexible,   % CRUCIAL: Fixes weird word spacing and alignment gaps!
    upquote=true            % Brings back the missing double quotes
  }
}
# Task
You are a Decontextualization Judge. Your task is to evaluate if a Decomposed Fact is **fully self-contained** and resolves all references to the source text.

## Input Description
You will be provided with two main inputs:
1. Context Information: A combined text block containing the original Source Text and optionally the User Question/Prompt. This is the ground truth for context.
2. Decomposed Fact: The decomposed fact (can be a sentence, triple, or triple with qualifiers) that needs to be evaluated for independence. Consider the entire fact including qualifiers, which contextualize the triple.

## Steps to Follow

### Step 1: The "Naive Reader" Scan
Read *only* the Decomposed Fact. Identify any words that act as "pointers" requiring external context to understand. Look specifically for:
- Pronouns (He, She, It, They, This, That).
- Definite References ("The study", "The decision" --- *Which one?*).
    - **Exception (Common Knowledge)**: References to widely known global entities or constants ("The Internet", "The Moon") do *not* count as pointers unless the context implies a specific, unique instance (e.g., "The President" of a local club).
- Relative Time/Place ("Next year", "Currently", "Locally").
- Comparative/Exclusionary Terms ("other", "another", "the remaining", "similar", "better", "such"). For example, "Other conferences" implies a specific set is excluded, and "Similar results" implies a comparison to a specific previous result.

### Step 2: The Source Availability Check
For every pointer identified in Step 1, look at the Context Information. Does the source contain the specific name, date, or list that the pointer refers to?
- *Example:* If the Decomposed Fact says "other conferences" and the Source says "She attended X and Y, and other conferences," the specific context (excluding X and Y) is available in the source.

### Step 3: Determine Status
Based on the availability of context, assign one of the following statuses:
- **FAIL (Resolvable)**: The Decomposed Fact uses a pointer, and the Source Text clearly defines what that pointer refers to. The Decomposed Fact should have included this info.
- **PASS (Inherited Ambiguity)**: The Decomposed Fact uses a pointer, but the Source Text is also vague (e.g., the source says "other places" without ever listing what the first places were).
- **PASS (Clear)**: The Decomposed Fact contains no unresolved pointers.

## Output Format
Provide a JSON response with the following structure:
```json
{
  "pointers_found": ["List of specific words like 'other conferences', 'next year'"],
  "source_context_available": "String explaining availability (e.g., 'Yes - Source mentions Youth for Climate')",
  "status": "PASS or FAIL"
}
```

### Input Data
Context Information:
$context

Decomposed Fact:
$decomposed_fact

### Output
\end{tcblisting}
\subsubsection{Prompt for qualifier check}
\label{sec:qualifier_misuse_check}
\begin{tcblisting}{
  colback=gray!5,
  colframe=gray!50!black,
  listing only,
  breakable,
  size=fbox, % Reduces gray padding
  title=Prompt for Qualifier Check,
  title after break={Prompt for Qualifier Check (Continued)},
  listing options={
    %basicstyle=\ttfamily\fontsize{8pt}{8.5pt}\selectfont,
    basicstyle=\ttfamily\scriptsize\linespread{0.8}\selectfont,
    breaklines=true,
    breakindent=0pt,        % Forces wrapped line indent to 0
    breakautoindent=false,  % Turns off the code-continuation indent
    columns=fullflexible,   % CRUCIAL: Fixes weird word spacing and alignment gaps!
    upquote=true            % Brings back the missing double quotes
  }
}
# Task: Qualifier Misuse Judge
Your task is to evaluate whether the qualifiers in decomposed hyper-relational facts are used correctly. You must check their structural application, semantic appropriateness, and contextual relevance against the original Target Text and the qualifier usage rules.

## STRICT CONSTRAINTS (What NOT to do)
- **NOT an entailment task:** Do not judge whether the fact is true, supported by the context, or hallucinated.
- **NOT a decontextualization task:** Do not judge whether the fact is self-contained or has unresolved pronouns.
- **NOT a coverage task:** Do not penalize missing facts or missing qualifiers.
- **Judge ONLY qualifier attachment:** Your sole focus is whether the existing qualifiers are incorrectly attached or labeled relative to the base triple and Target Text.

## Decomposed Fact Representation Formats:
- **Relation Triplet (Simple Facts)**:
    - Use this for standalone facts that do not require additional context.
      ```json
      {"base_triple": ["subject", "relation", "object"]}
      ```
- **Hyper-Relational Tuple (Base Triple + Qualifiers)**:
    - Use this when additional details (e.g., context, causality, sequence, provenance) are needed to clarify the base triplet.
      ```json
      {"base_triple": ["subject", "relation", "object"],
         "qualifiers": [{"qualifier-label-1": "qualifier-value-1"}, {"qualifier-label-2": "qualifier-value-2"}]}
      ```

Note: This evaluation receives only hyper-relational facts with at least one qualifier. Simple facts are not evaluated here. They matter only as the cleaner alternative when a qualifier expresses an independent background fact about the subject or object.

## Valid Qualifier Functions:
Because this is Open Information Extraction, do not enforce a strict ontology. Natural-language labels, generic prepositions, and labels copied from the source text are acceptable if they preserve the semantic role.

A valid qualifier directly contextualizes, refines, or specifies the base triple. Valid qualifiers commonly express:
- validity context: time, location, condition, scope;
- causality/purpose: cause, reason, purpose;
- sequence/order: before, after, replaces;
- provenance/method: source, method, instrument, according to;
- role/relationship: as, with, by, shared with;
- local description/annotation/constraint: type, category, amount, score, exception, despite.


## Qualifier Usage Rules:
### Rule 1: Modify the Base Triple Only
A qualifier must directly modify or clarify the specific Subject-Relation-Object assertion. It must not modify another qualifier.

### Rule 2: Local Relation-Relevance
A qualifier may describe the subject, object, event, or another locally mentioned item. This is valid if the qualifier helps preserve how the Target Text frames the base triple.

Ask:
- Is the qualifier locally tied to this base triple in the Target Text?
- Does it help specify, disambiguate, contrast, add context to, or constrain the subject/object/relation in this specific claim?
- Or is it an unrelated background fact that would be equally natural even if this base triple were removed?

Do **not** flag a qualifier merely because it describes the subject or object, or because it could also be extracted as a separate simple fact. Many valid qualifiers describe a local type, domain, role, contrast, additive relation, or category.

Accept qualifiers such as:
- object type/category: `brand type: international brand` in "campaigns for international brands such as L'Oreal";
- domain: `sport: rugby league` in "represented England at under-18 level" when the text is about rugby league;
- additive/contrastive framing: `in addition to: The Crown` in "appeared in Mission: Impossible - Fallout in addition to The Crown";
- role: `role: Cobb` in "starred as Cobb";
- local scope/category: `public` in "public universities offer tuition-free education".

Flag `Decomposition Error (Subject Attribute)` or `Decomposition Error (Object Attribute)` only when the qualifier is an unrelated independent background property that does not help interpret the base triple and is not locally used to frame the claim.

If a qualifier is locally meaningful but structurally imperfect, mark `qualifier_misuse_detected` as `"No"`.

### Rule 3: Semantic Role Preservation
The qualifier label must reasonably describe the semantic role between the base triple and the qualifier value.

Do not penalize minor label imprecision. For example, `time` vs. `point in time`, or generic labels like `for`, `with`, `as`, and `by`, are acceptable if the intended semantic role is clear.

Broad labels are acceptable only when the intended attachment is clear from the base triple and Target Text. If a broad label such as `attribute`, `description`, `type`, or `status` makes it unclear whether the qualifier describes the subject, object, or relation, and this ambiguity could change the meaning or make verification unreliable, mark `Semantic Label Error`.

Only flag `Semantic Label Error` if the qualifier label fundamentally changes the meaning of the Target Text or creates harmful attachment ambiguity.

## Misuse Types
Mark `qualifier_misuse_detected` as "Yes" only if at least one qualifier has one of these errors:
- `Structural Error`: The qualifier modifies another qualifier rather than the base triple.
- `Decomposition Error (Subject Attribute)`: The qualifier gives an unrelated independent background property of the subject, rather than locally contextualizing how the subject participates in the base triple.
- `Decomposition Error (Object Attribute)`: The qualifier gives an unrelated independent background property of the object, rather than locally contextualizing how the object participates in the base triple.
- `Semantic Label Error`: The qualifier label fundamentally misrepresents the semantic role, or is so ambiguous that it becomes unclear whether the qualifier modifies the subject, object, or base relation.

## Output Schema
Important: Only include the `error_type` key if `qualifier_misuse_detected` is "Yes". Omit it entirely if the value is "No".
Output exactly one JSON object and no extra text.

Use this schema:
```json
{
  "reasoning_process": "Concise explanation focused only on qualifier usage.",
  "qualifier_misuse_detected": "Yes or No",
  "error_type": "Structural Error | Decomposition Error (Subject Attribute) | Decomposition Error (Object Attribute) | Semantic Label Error"
}
```
If multiple errors exist, output the error_type of the clearest.

## Examples:
### Example 1: Valid Object Description:
#### Context:
A fashion model's commercial work was discussed in a biography.
#### Target Text:
Jessica Barboza has also been featured in advertising campaigns for international brands such as L'Oreal, Garnier, and Maybelline.

#### Decomposed Fact:
```json
{
  "base_triple": ["Jessica Barboza", "has been featured in advertising campaigns for", "L'Oreal"],
  "qualifiers": [{"brand type": "international brand"}]
}
```

#### Evaluation:
```json
{
  "reasoning_process": "The qualifier 'brand type: international brand' describes L'Oreal as part of the local phrase 'international brands such as L'Oreal'. It helps specify the object in this campaign relation and preserves the source meaning.",
  "qualifier_misuse_detected": "No"
}
```

### Example 2: Decomposition Error (Object Attribute)
#### Context:
Many fans were eager to learn more about the cast's background.
#### Target Text:
'Inception' starred Leonardo DiCaprio, who was born in November 1974.

#### Decomposed Fact:
```json
{"base_triple": ["Inception", "starring", "Leonardo DiCaprio"],
 "qualifiers": [{"date of birth": "November 1974"}]}
```

#### Evaluation:
```json
{
  "reasoning_process": "The qualifier 'date of birth: November 1974' is an independent background property of Leonardo DiCaprio. It does not specify his participation in the starring relation and should be a separate simple fact.",
  "qualifier_misuse_detected": "Yes",
  "error_type": "Decomposition Error (Object Attribute)"
}
```

### Example 3: Valid Subject Description
#### Context:
An international dance competition announced its category winners.
#### Target Text:
French choreographer Camille Laurent represented France in the contemporary dance category at the 2024 Lyon Dance Biennale.

#### Decomposed Fact:
```json
{
  "base_triple": ["Camille Laurent", "represented", "France"],
  "qualifiers": [{"profession": "choreographer"}, {"category": "contemporary dance"}, {"event": "2024 Lyon Dance Biennale"}]
}
```

#### Evaluation:
```json
{
  "reasoning_process": "Although 'profession: choreographer' describes the subject, it is locally relevant because it helps specify the artistic capacity in which Camille Laurent represented France. The category and event qualifiers also directly contextualize the representation relation.",
  "qualifier_misuse_detected": "No"
}
```

### Example 4: Decomposition Error (Subject Attribute)
#### Context:
The history of science is filled with pioneering figures.
#### Target Text:
Marie Curie, who was born in Warsaw, was awarded the Nobel Prize in Physics in 1903.

#### Decomposed Fact:
```json
{"base_triple": ["Marie Curie", "received award", "Nobel Prize in Physics"],
 "qualifiers": [{"point in time": "1903"}, {"place of birth": "Warsaw"}]}
```

#### Evaluation:
```json
{
  "reasoning_process": "The time qualifier directly contextualizes the award event, but 'place of birth: Warsaw' is an independent background property of Marie Curie and does not contextualize receiving the award.",
  "qualifier_misuse_detected": "Yes",
  "error_type": "Decomposition Error (Subject Attribute)"
}
```

### Example 5: Semantic Label Error (Meaning-Changing Label)
#### Context:
The annual medical conference concluded with a highly anticipated presentation on cardiovascular health. Hundreds of attendees gathered in the main auditorium.
#### Target Text:
The study's conclusion was presented by Dr. Anya Sharma on behalf of the National Medical Institute.

#### Decomposed Fact:
```json
{"base_triple": ["Study's conclusion", "presented by", "Dr. Anya Sharma"],
 "qualifiers": [{"role": "National Medical Institute"}]}
```

#### Evaluation:
```json
{
  "reasoning_process": "The qualifier value 'National Medical Institute' is the organization on whose behalf the presentation was made, not a role. The label 'role' fundamentally misrepresents the semantic relation.",
  "qualifier_misuse_detected": "Yes",
  "error_type": "Semantic Label Error"
}
```

### Example 6: Semantic Label Error (Ambiguous Qualifier Attachment)
#### Context:
A charity concert featured several well-known singers and songs.
#### Target Text:
Singer Maya Lee performed the popular ballad "Silver Rain" at the charity concert.

#### Decomposed Fact:
```json
{
  "base_triple": ["Maya Lee", "performed", "Silver Rain"],
  "qualifiers": [{"attribute": "popular"}]
}
```

#### Evaluation:
```json
{
  "reasoning_process": "The Target Text describes the ballad 'Silver Rain' as popular, but the broad label 'attribute' makes it unclear whether 'popular' describes Maya Lee, the song, or the performance. The qualifier should use a clearer label such as 'song attribute'.",
  "qualifier_misuse_detected": "Yes",
  "error_type": "Semantic Label Error"
}
```

### Example 7: Structural Error
#### Context:
A technology company issued a press release before a product event.
#### Target Text:
Apple announced its new product line according to a press release published on Monday.

#### Decomposed Fact:
```json
{
  "base_triple": ["Apple", "announced", "new product line"],
  "qualifiers": [{"according to": "a press release"}, {"publication date": "Monday"}]
}
```

#### Evaluation:
```json
{
  "reasoning_process": "The qualifier 'according to: a press release' directly contextualizes the announcement, but 'publication date: Monday' describes the press release rather than directly modifying the base triple about Apple announcing the product line. It should be represented as a separate fact about the press release.",
  "qualifier_misuse_detected": "Yes",
  "error_type": "Structural Error"
}
```

### Example 8: Valid Qualifier Usage
#### Context:
The acclaimed director had a phenomenal year in 2019.
#### Target Text:
Taika Waititi won the Academy Award for Best Adapted Screenplay in 2019 for Jojo Rabbit.
#### Decomposed Fact:
```json
{"base_triple": ["Taika Waititi", "won", "Academy Award for Best Adapted Screenplay"],
 "qualifiers": [{"time": "2019"}, {"for": "Jojo Rabbit"}]}
```

#### Evaluation:
```json
{
  "reasoning_process": "Both qualifiers directly contextualize the award event: 'time' gives when the award was won, and 'for' gives the work associated with the award. The generic labels preserve the semantic roles.",
  "qualifier_misuse_detected": "No"
}
```

## Input:
### Context:
$context
### Target Text:
$target_text

### Decomposed Fact:
$decomposed_fact

### Evaluation:
\end{tcblisting}
\twocolumn
\subsection{Experiment setup}
\label{sec:experiment_setup}
All experiments were conducted on a single compute node equipped with eight NVIDIA H200 GPUs. We utilized vLLM~\citep{kwon2023efficient} as our primary inference engine, integrated with LiteLLM's Batch Completion \footnote{\url{https://github.com/BerriAI/litellm}} feature to manage requests.

We enabled the reasoning/thinking mode across all three evaluated LLM families. Since current open-weight models often generate extensive thinking processes prior to outputting a final answer and are consequently prone to getting stuck in repetitive loops, we applied a generation temperature greater than 0.0. Specifically, we adopted the optimal temperature and sampling settings for reasoning tasks as recommended by each model's respective Hugging Face model card~\footnote{\url{https://huggingface.co/models}}.

The precise sampling parameters used for each model are summarized in Table~\ref{tab:model_params}:
\begin{table}[!htbp]
\small
\centering
\begin{tabular}{lccc}
\toprule
\textbf{Model} & \textbf{Temperature} & \textbf{Top-p} & \textbf{Top-k} \\ \midrule
Qwen3.5-397B-A17B & 0.6 & 0.95 & 20 \\
Qwen3.5-122B-A10B & 1.0 & 0.95 & 20 \\
Qwen3.5-35B-A3B   & 1.0 & 0.95 & 20 \\
GLM-5-FP8 (744B)      & 1.0 & 0.95 & 50 \\
GPT-OSS-120B      & 1.0 & 1.00   & 50 \\ \bottomrule
\end{tabular}
\caption{Sampling parameters for evaluated LLMs.}
\label{tab:model_params}
\end{table}

\subsection{Decomposition quality evaluation}
\subsubsection{Cohen's kappa for decomposition quality evaluation}
\label{sec:cohen_kappa}
Table~\ref{tab:judge_agreement} reports Cohen's kappa between the two judge models for the binary entailment and decontextualization checks. These values should be interpreted together with the label distributions in Table~\ref{tab:decomp_quality}, since kappa can be deflated under highly imbalanced labels. This is especially relevant for the entailment check on FactScore-Bio dataset, where both judges assign Entailment to the vast majority (over 99\%) of extracted facts.

\begin{table}[!htbp]
\centering
\scriptsize
\begin{tabular}{llcc}
\toprule
\textbf{Dataset} & \textbf{Framework} & \textbf{Entailment $\kappa$} & \textbf{Decontext. $\kappa$} \\
\midrule
\multirow{3}{*}{FactScore-Bio}
& RefChecker & 0.42 & 0.65 \\
& VeriScore  & 0.26 & 0.60 \\
& TriQua     & 0.48 & 0.69 \\
\midrule
\multirow{3}{*}{LongFact}
& RefChecker & 0.70 & 0.67 \\
& VeriScore  & 0.63 & 0.52 \\
& TriQua     & 0.54 & 0.52 \\
\bottomrule
\end{tabular}
\caption{
Inter-judge agreement between Qwen3.5-397B-A17B and GLM-5-FP8 for the automatic decomposition-quality checks.
Cohen's $\kappa$ is reported for binary entailment and decontextualization labels.
Because label distributions are highly imbalanced, $\kappa$ should be interpreted together with the entailment and decontextualization rates in Table~\ref{tab:decomp_quality}.
}
\label{tab:judge_agreement}
\end{table}

\subsubsection{Qualifier usage error rate}
\label{sec:error_rate}
Table~\ref{tab:qualifier_separate_error_rates} breaks down qualifier-misuse errors. Relative percentages are computed over detected misuse cases. The bottom row reports the overall misuse rate over all qualified TriQua facts. 

The two judges agree on low overall error rates but differ in category assignment. FactScore-Bio errors are mostly semantic-label and subject-attribute cases. LongFact shows more structural errors, likely due to longer and more abstract input sentences. We use error types only for qualitative analysis.
\begin{table}[!htbp]
    \centering
    \resizebox{\columnwidth}{!}{%
    \begin{tabular}{l cc cc}
        \toprule
        & \multicolumn{2}{c}{\textbf{FactScore-Bio}} & \multicolumn{2}{c}{\textbf{LongFact}} \\
        \cmidrule(lr){2-3} \cmidrule(l){4-5}
        \textbf{Error Type (Relative \%)} & \textbf{Qwen3.5} & \textbf{GLM-5} & \textbf{Qwen3.5} & \textbf{GLM-5} \\
        \midrule
        Decomposition (Subject Attribute) & 47.62 & 23.53 & 33.33 & 6.35 \\
        Decomposition (Object Attribute)  & 0 & 0 & 6.90 & 0 \\
        Structural Error                  & 4.76 & 0 & 33.33 & 60.32 \\
        Semantic Label Error              & 47.62 & 76.47 & 26.44 & 33.33 \\
        \midrule
        \textbf{Overall Error Rate (\%)}  & 1.34 & 1.08 & 2.22 & 1.61 \\
        \bottomrule
    \end{tabular}%
    }
    \caption{Relative distribution of qualifier misuse error types (as a percentage of total errors) and the overall qualifier error rate across both datasets.}
    \label{tab:qualifier_separate_error_rates}
\end{table}

\onecolumn
\subsection{CLEARFACTS}
\subsubsection{CLEARFACTS subsets statistics}
\label{sec:clearfacts_subsets_stats}
Table~\ref{tab:clearfacts_stats} shows the sample sizes and average token numbers of each subset in the CLEARFACTS dataset.

\begin{table}[!htbp]
\centering
\small
\begin{tabular}{lcc}
\toprule
\textbf{Subset} & \textbf{Sample Size (N)} & \textbf{Avg. Tokens} \\
\midrule
AggreFact   & 982 & 31.22 \\
SciFact     & 89  & 20.72 \\
Cover  & 238 & 28.55 \\
Hover       & 281 & 35.89 \\
\bottomrule
\end{tabular}
\caption{Statistics of CLEARFACT subsets. The mean token length of claims in each subset is measured with Qwen3.5-397B-A17B as tokenizer.}
\label{tab:clearfacts_stats}
\end{table}

\subsubsection{CLEARFACTS full results}
\label{sec:clearfacts_result_full}
Table~\ref{tab:clearfacts_results_full} shows verification results on CLEARFACTS dataset across all backbone models.

\begin{table*}[!htbp]
\centering
\footnotesize 
\setlength{\tabcolsep}{4.5pt} 
\begin{tabular}{llccccc}
\toprule
\multirow{2}{*}{\textbf{Backbone Model}} & \multirow{2}{*}{\textbf{Framework}} & \multicolumn{4}{c}{\textbf{Subset F1 (\%)}} & \multirow{2}{*}{\textbf{Overall}} \\
\cmidrule(lr){3-6}
& & \textbf{AggreFact} & \textbf{SciFact} & \textbf{Cover} & \textbf{Hover} & \\
\midrule
\multirow{4}{*}{Qwen3.5-397B-A17B} 
& Few-Shot (No Decomp.) & 88.30 & 89.03 & 78.99 & 83.89 & 86.22 \\
\cmidrule{2-7}
& RefChecker & 82.89 & 85.39 & 73.9 & 77.65 & 81.11 \\
& VeriScore & 84.60 & \textbf{87.41} & 78.84 & \textbf{83.55} & 83.63 \\
& \textbf{TriQua (Ours)} & \textbf{87.98} & 87.23 & \textbf{80.02} & 82.79 & \textbf{85.91} \\
\midrule

\multirow{4}{*}{Qwen3.5-122B-A10B} 
& Few-Shot (No Decomp.) & 89.60 & 89.02 & 81.28 & 85.20 & 87.59 \\
\cmidrule{2-7}
& RefChecker & 83.91  & 85.98 & 72.59 & 78.59 & 81.49 \\
& VeriScore & 83.14 & \textbf{90.29} & 73.85 & \textbf{83.42} & 82.53 \\
& \textbf{TriQua (Ours)} & \textbf{85.56} & 85.37 & \textbf{77.94} & 79.58 & \textbf{83.46} \\
\midrule
\multirow{4}{*}{Qwen3.5-35B-A3B} 
& Few-Shot (No Decomp.) & 88.05  & 87.23 & 80.04 & 84.44 & 86.22 \\
\cmidrule{2-7}
& RefChecker & 81.01  & 86.06 & 68.53 & 74.76 & 78.42 \\
& VeriScore & 82.85 & 88.06 & \textbf{77.12} & \textbf{81.46} & 82.30 \\
& \textbf{TriQua (Ours)} & \textbf{84.66} & \textbf{89.03} & 76.88 & 78.22 & \textbf{82.70} \\
\midrule
\multirow{4}{*}{GLM-5-FP8 (744B)} 
& Few-Shot (No Decomp.) & 85.23 & 82.24 & 78.14 & 83.32 & 83.74 \\
\cmidrule{2-7}
& RefChecker & 81.16 & 84.14 & 70.49 & 73.67 & 78.53 \\
& VeriScore & 82.81 & \textbf{88.48} & 73.61 & \textbf{81.18} & 81.59 \\
& \textbf{TriQua (Ours)} & \textbf{83.24} & 86.57 & \textbf{78.14} & 78.08 & \textbf{81.85} \\
\midrule
\multirow{4}{*}{GPT-OSS-120B} 
& Few-Shot (No Decomp.) & 88.54 & 90.22 & 77.51 & 83.93 & 86.23 \\
\cmidrule{2-7}
& RefChecker & 78.75  & 85.37 & 65.61 & 73.00 & 76.26 \\
& VeriScore & 82.89 & \textbf{89.25} & \textbf{75.75} & \textbf{82.34} & 82.23 \\
& \textbf{TriQua (Ours)} & \textbf{84.29} & 87.83 & 75.21 & 79.28 & \textbf{82.33} \\
\bottomrule
\end{tabular}
\caption{F1 scores on the CLEARFACTS dataset across various backbone models. In all decomposition frameworks, the specified backbone model functions as both decomposer and verifier. TriQua is benchmarked against alternative decomposition pipelines and a direct-verification (no decomposition) baseline. Results are averaged across two independent runs for each setup.}
\label{tab:clearfacts_results_full}
\end{table*}

\subsubsection{Comparison of TriQua and Few-Shot (No Decomp.) on the CLEARFACTS dataset}
\label{sec:comparision_direct_triqua}
Table~\ref{tab:clearfacts_comparison_no_decomp} provides a detailed comparison between TriQua and few-shot direct verification without decomposition. The main performance gap comes from recall on the Supported class. TriQua is more conservative and more often predicts Not Supported for examples labeled as Supported in the gold data.

\begin{table*}[!htbp]
\centering
\small
\begin{tabular}{llrrrrr}
\toprule
\textbf{Backbone} & \textbf{Method} & \textbf{Macro F1} & \textbf{S Prec.} & \textbf{S Rec.} & \textbf{NS Prec.} & \textbf{NS Rec.} \\
\midrule
\multirow{2}{*}{Qwen3.5-397B-A17B}
& Few-Shot (No Decomp.) & \textbf{86.22} & \textbf{97.48} & 77.59 & 77.06 & \textbf{97.40} \\
& TriQua     & 85.91 & 96.35 & \textbf{77.98} & \textbf{77.14} & 96.18 \\
\midrule
\multirow{2}{*}{Qwen3.5-122B-A10B}
& Few-Shot (No Decomp.) & \textbf{87.59} & \textbf{96.67} & \textbf{80.82} & \textbf{79.52} & 96.39 \\
& TriQua     & 83.46 & 96.48 & 73.36 & 73.69 & \textbf{96.54} \\
\midrule
\multirow{2}{*}{Qwen3.5-35B-A3B}
& Few-Shot (No Decomp.) & \textbf{86.22} & \textbf{97.55} & \textbf{77.54} & \textbf{77.02} & \textbf{97.47} \\
& TriQua     & 82.70 & 96.77 & 71.74 & 72.59 & 96.90 \\
\midrule
\multirow{2}{*}{GLM-5-FP8}
&Few-Shot (No Decomp.) & \textbf{83.74} & \textbf{97.69} & \textbf{72.91} & \textbf{73.60} & \textbf{97.76} \\
& TriQua     & 81.85 & 95.29 & 70.85 & 71.81 & 96.10 \\
\midrule
\multirow{2}{*}{GPT-OSS-120B}
& Few-Shot (No Decomp.) & \textbf{86.23} & 94.38 & \textbf{80.43} & \textbf{78.74} & 93.80 \\
& TriQua     & 82.33 & \textbf{94.70} & 72.74 & 72.86 & \textbf{94.73} \\
\bottomrule
\end{tabular}
\caption{
CLEARFACTS stress-test results comparing direct verification without decomposition and TriQua.
All values are percentages. S and NS denote Supported and Not Supported labels, respectively.
Best values within each backbone block are bolded.
}
\label{tab:clearfacts_comparison_no_decomp}
\end{table*}

\subsubsection{TriQua qualitative error analysis}
\label{sec:triqua_qualitative_error_analysis}
Table~\ref{tab:clearfacts_triqua_error_analysis} categorizes 50 sampled cases in which the gold label and direct verification are Supported, while TriQua predicts Not Supported. The two largest sources are flawed original claims or gold labels, and TriQua over-strictness. Although CLEARFACTS applies a dedicated filtering pipeline to improve annotation quality, its authors acknowledge that residual annotation errors may remain~\citep{seo2025verifying}. In our manual inspection, such cases occur particularly often in the AggreFact subset, which also constitutes the largest portion (62\%) of CLEARFACTS. Similar annotation issues for AggreFact have also been reported by \citet{laban-etal-2023-summedits}.

The second major source is TriQua's conservative component-level verification. Unlike holistic direct verification, TriQua verifies each base triple and attached qualifier separately. Consequently, one unsupported or over-strictly judged component can make the aggregated claim prediction Not Supported.

\begin{table*}[!htbp]
\centering
\small
\begin{tabular}{p{0.30\linewidth}rp{0.56\linewidth}}
\toprule
\textbf{Error Source} & \textbf{Count} & \textbf{Explanation} \\
\midrule
Original claim flawed & 23 &
The original claim contains an overclaim, wrong scope, unsupported modifier, or malformed wording. Direct verification accepts it holistically, while TriQua flags the unsupported component. \\
\midrule
TriQua verification error / over-strictness & 20 &
The evidence supports the claim under reasonable paraphrase, inference, or context, but TriQua rejects it as neutral/not-supported under its evidence-only policy. \\
\midrule
TriQua decomposition or representation error & 4 &
The decomposition changes the scope, drops context, or attaches a qualifier/modifier to the wrong argument. \\
\midrule
Evidence or claim artifact & 2 &
The disagreement is caused by a typo or source-internal inconsistency. \\
\midrule
Task-boundary issue & 1 &
The input is an imperative or non-factual instruction and is not well suited to factual triple extraction and verification. \\
\bottomrule
\end{tabular}
\caption{
Manual analysis of 50 CLEARFACTS cases where the gold label and direct verification predict Supported, while TriQua predicts Not Supported.
}
\label{tab:clearfacts_triqua_error_analysis}
\end{table*}

\end{document}